\documentclass{article}

\usepackage[preprint]{neurips_2024}

\usepackage[utf8]{inputenc}
\usepackage[T1]{fontenc}
\usepackage{hyperref}
\usepackage{url}
\usepackage{booktabs}
\usepackage{amsfonts}
\usepackage{nicefrac}
\usepackage{microtype}
\usepackage{xcolor}
\usepackage{graphicx}
\usepackage{subcaption}
\usepackage{amsmath}
\usepackage{amssymb}
\usepackage{float}
\usepackage{longtable}

\title{SpatialBench: Can Agents Analyze Real-World Spatial Biology Data?}

\author{
  Kenny Workman \hspace{1.5em}
  Zhen Yang \hspace{1.5em}
  Harihara Muralidharan \hspace{1.5em}
  Hannah Le \\[1em]
  LatchBio, San Francisco, CA \\[0.5em]
  Correspondence: \texttt{kenny@latch.bio}
}

\begin{document}
\raggedbottom

\maketitle

\begin{abstract}
Spatial transcriptomics assays are rapidly increasing in scale and complexity,
making computational analysis a major bottleneck in biological discovery.
Although frontier AI agents have improved dramatically at software engineering
and general data analysis, it remains unclear whether they can extract
biological insight from messy, real-world spatial datasets. We introduce
SpatialBench, a benchmark of 146 verifiable problems derived from practical
spatial analysis workflows spanning five spatial technologies and seven task
categories. Each problem provides a snapshot of experimental data immediately
prior to an analysis step and a deterministic grader that evaluates recovery of
a key biological result. Benchmark data on frontier models shows that base
model accuracy remains low (20-38\% across model families), with strong
model--task and model--platform interactions. Harness design has a large
empirical effect on performance, indicating that tools, prompts, control flow,
and execution environment should be evaluated and improved as first-class
objects. SpatialBench serves both as a measurement tool and a diagnostic lens
for developing agents that can interact with real spatial datasets faithfully,
transparently, and reproducibly.
\end{abstract}

\section{Introduction}
\label{sec:introduction}

Modern biology experiments increasingly rely on spatial assays: data-intensive
measurement tools that capture molecular state \textit{in situ} to enable the
study of disease, development, and tissue organization within their native
spatial context~\citep{stahl2016spatial,chen2015merfish,moses2022museum,liu2024spatiotemporal}. The raw outputs of these assays are rapidly growing in both
scale and complexity~\citep{moses2022museum,liu2024spatiotemporal}, making it difficult for classically trained biologists to
draw scientific conclusions without skill in programming and data analysis. As
a result, unanalyzed spatial data is creating a growing bottleneck in
biological discovery.

Over the past year, AI agents have advanced rapidly in software engineering and
general data analysis~\citep{jimenez2024swebench,yang2024sweagent,yao2023react}. Biology-specific analogues are now beginning to emerge,
with the promise of enabling scientists to drive complex analyses using
natural-language instructions, allowing questions to be answered directly from
raw data without requiring procedural coding or data analysis skills~\citep{yao2023react,schick2023toolformer}. 

However, in their current form, agents for biological data remain both
unreliable and underpowered. They are prone to scientific inaccuracies and
misleading behavior (including hallucinations)~\citep{huang2023hallucination_survey,singhal2023multimedqa}, and frequently fail to complete
domain-specific analysis tasks unique to spatial workflows. Quantitative
benchmarks will therefore be essential for measuring progress, comparing
systems, and guiding model improvement. Yet existing evaluations focus
primarily on general biological knowledge~\citep{tsatsaronis2015bioasq,jin2019pubmedqa,singhal2023multimedqa} and are not representative of real-world
tasks that manipulate and draw conclusions from messy data.

We introduce SpatialBench, a suite of 146 verifiable problems distilled from
real spatial analysis workflows, where each item snapshots an analysis state
immediately before a target step and is paired with a deterministic grader. We
quantify how frontier models behave on messy spatial data—revealing strong
model--task and model--platform interactions—and show that harness design
(tools, prompts, control flow, execution environment) can change outcomes as
much as the choice of base model. SpatialBench is both a yardstick to measure
progress and a diagnostic tool to scaffold test-driven development of agents
that analyze real spatial datasets faithfully, transparently, and reproducibly.

\section{Results}
\label{sec:results}

\subsection{SpatialBench is a Suite of Verifiable Problems Constructed from
Real Workflows}
\label{sec:benchmark-construction}

\begin{figure}[ht]
  \centering
  \includegraphics[width=\textwidth]{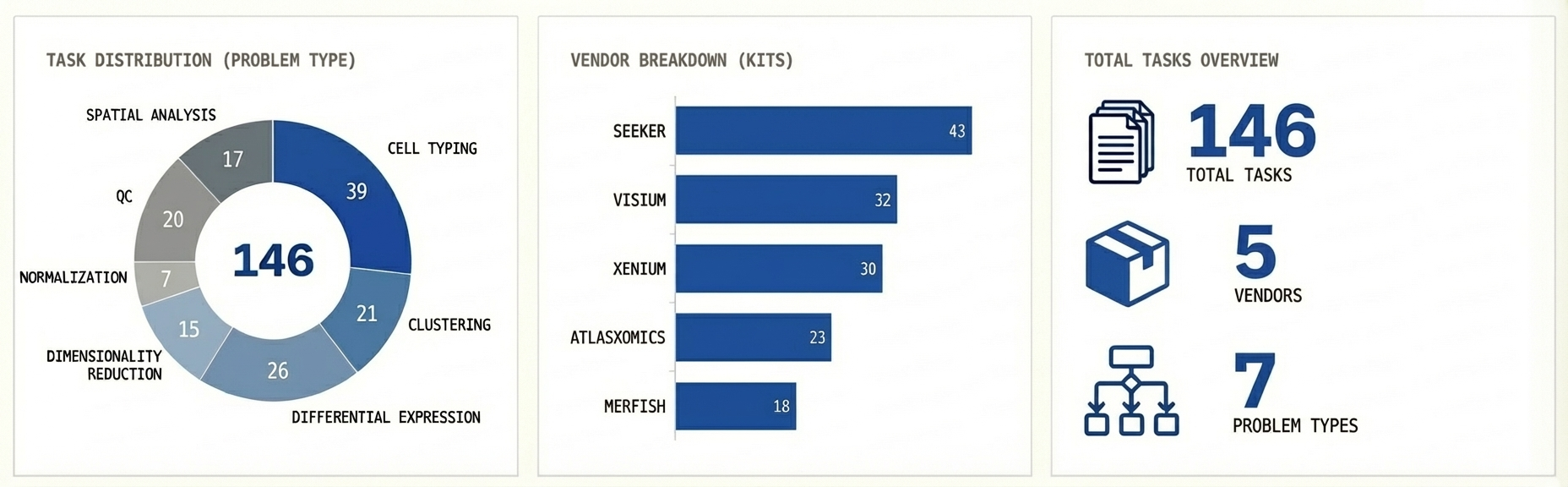}
  \caption{Overview of SpatialBench.}
  \label{fig:benchmark-overview}
\end{figure}

To construct a benchmark that approximates real-world tasks in spatial data
analysis (Figure~\ref{fig:benchmark-overview}), we collaborated with scientists
and spatial technology manufacturers across diverse tissue types and disease
contexts. We decomposed end-to-end analysis workflows into gradeable steps,
such as quality control, normalization, and cell typing. At each step, we
sought to extract the key biological idea and formalize tacit and artisanal
pattern recognition into deterministic heuristics.

\begin{figure}[ht]
  \centering
  \includegraphics[width=\textwidth]{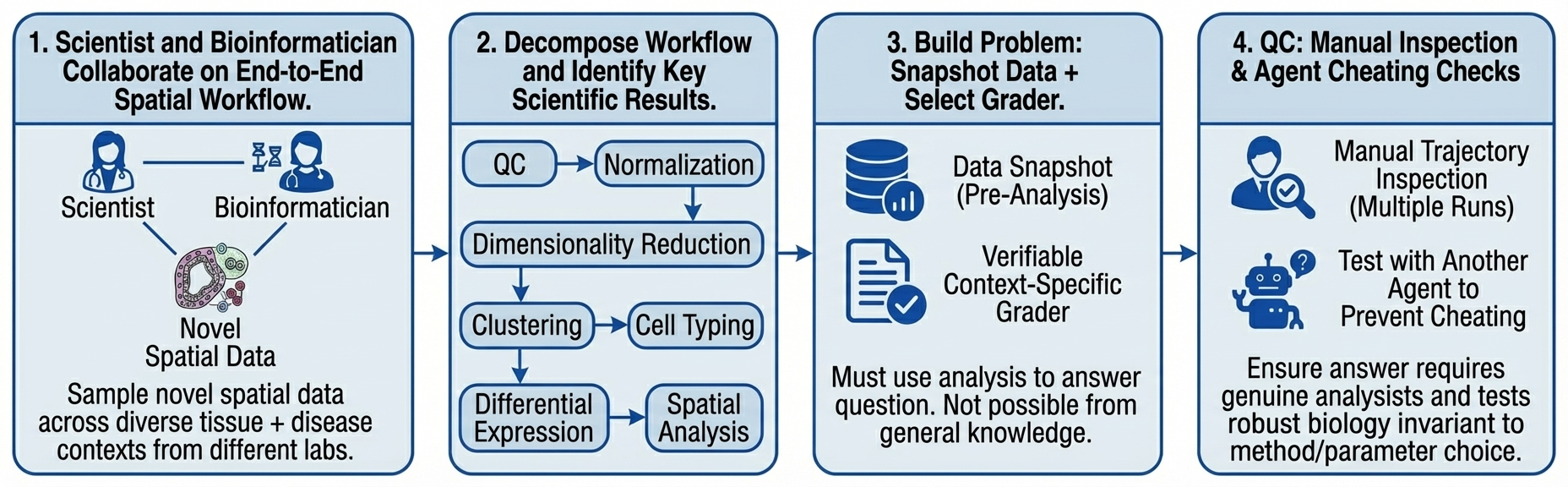}
  \caption{SpatialBench benchmark construction.}
  \label{fig:benchmark-construction}
\end{figure}

The final evaluation suite consists of 146 problems spanning five spatial
transcriptomics technologies and seven task categories
(Figure~\ref{fig:benchmark-construction}). Each problem includes a snapshot of
real experimental data taken immediately prior to a target analysis step, a
description of the task through a high-level scientific lens, and a deterministic
grader (e.g., Jaccard similarity of sets) that evaluates recovery of the key
biological result in a verifiable manner.

To ensure robustness, all problems underwent manual quality control, including
inspection of agent trajectories across multiple runs and adversarial testing
against agents instructed to exploit shortcuts or prior knowledge.

The benchmark is designed to test durable biological reasoning rather than
method-specific implementation details.

Correct answers are generally robust to reasonable choices of algorithms and
hyperparameters. For example, early principal components in dimensionality
reduction tasks are expected to separate broad biological programs that are
specific to the dataset yet invariant to reasonable preprocessing choices. A
good SpatialBench problem therefore tests scientific interpretation of these
stable patterns; for instance, asking which of two cell populations is more
strongly separated along the first principal component.

In addition, tasks explicitly require empirical interaction with the data.
Agents that rely on prior biological knowledge or pattern matching without
performing the requisite data manipulation and inspection fail to complete many
tasks correctly, even when the underlying biological concepts are well known.

All results are reported over triplicate runs~\citep{jimenez2024swebench}, with confidence intervals
constructed by aggregating uncertainty over pooled per-task mean scores.

\subsection{SpatialBench Stratifies Frontier Models on Performance and Efficiency}
\label{sec:aggregate-results}

\begin{figure}[H]
  \centering
  \includegraphics[width=\textwidth]{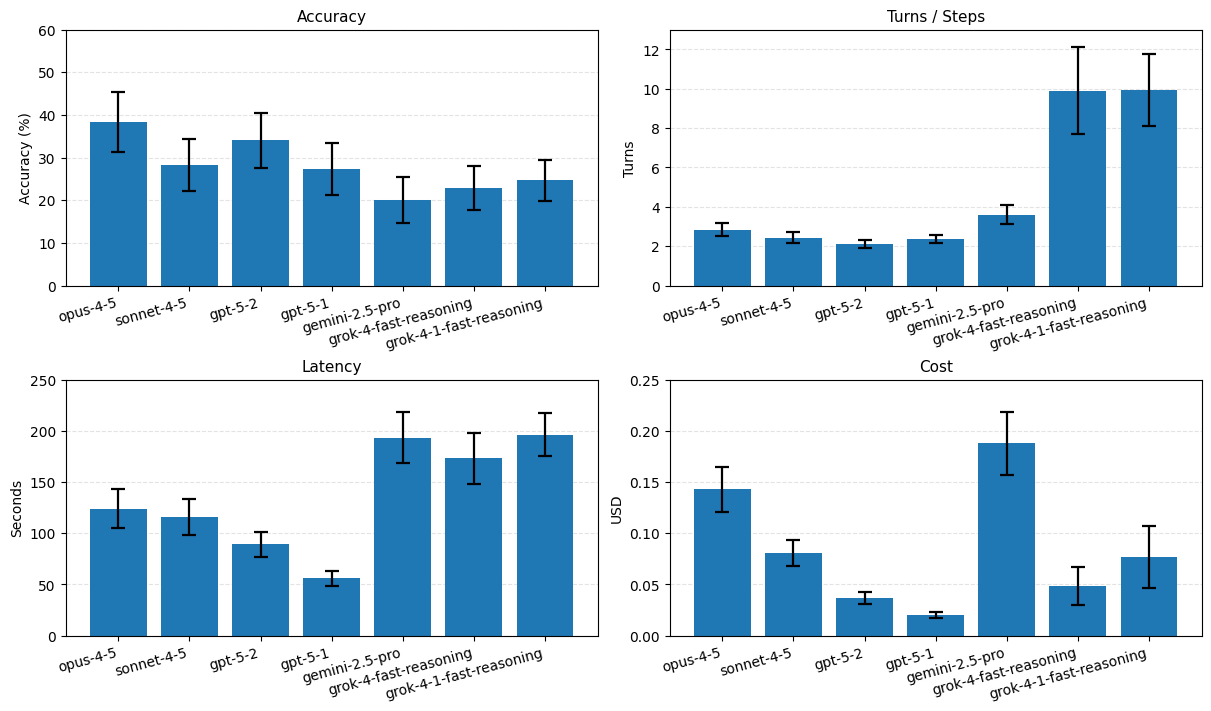}
  \caption{Aggregate model performance across SpatialBench.}
  \label{fig:aggregate-performance}
\end{figure}

Across the full benchmark (Figure~\ref{fig:aggregate-performance}, Table~\ref{tab:aggregate}),
Opus-4.5 attains the highest mean accuracy (38.4\%),
followed by GPT-5.2 (34.0\%) and Sonnet-4.5 (28.3\%), while Gemini-2.5-Pro
(20.1\%) and Grok variants (22--25\%) consistently underperform. In contrast to
accuracy, efficiency metrics exhibit substantially stronger separation across
model families. GPT-5.1 and GPT-5.2 operate at markedly lower cost
(approximately \$0.02--\$0.04 per evaluation) and lower latency (56--89\,s)
than Anthropic, Gemini, and Grok models, whose costs exceed \$0.08 and whose
latencies typically exceed 115\,s. Differences in control flow are even more
pronounced: GPT and Anthropic models complete tasks in approximately 2--3 steps
on average, whereas Grok variants require nearly fourfold more steps ($\sim$10)
and frequently approach the maximum step budget, coinciding with the
highest latencies observed. As a result, differences in efficiency metrics
dominate their associated uncertainty, yielding a clear trade-off between peak
accuracy (Opus-4.5) and cost-effective performance (GPT-5.2).

\begin{table}[H]
\centering
\caption{Aggregate performance and efficiency across the full SpatialBench benchmark.
Reported values are mean $\pm$ 95\% confidence intervals.}
\label{tab:aggregate}
\begin{tabular}{lcccc}
\toprule
Model & Accuracy (\%) & Steps & Latency (s) & Cost (USD) \\
\midrule
Opus-4.5        & 38.36 [31.27, 45.44] & 2.84 [2.51, 3.17] & 123.8 [104.8, 142.9] & 0.143 [0.121, 0.165] \\
Sonnet-4.5      & 28.31 [22.22, 34.40] & 2.43 [2.17, 2.70] & 115.6 [98.0, 133.2]  & 0.081 [0.068, 0.093] \\
GPT-5.2         & 34.02 [27.57, 40.47] & 2.10 [1.89, 2.30] & 89.2  [76.9, 101.4]  & 0.037 [0.031, 0.042] \\
GPT-5.1         & 27.40 [21.30, 33.50] & 2.38 [2.18, 2.58] & 55.8  [48.2, 63.5]   & 0.020 [0.017, 0.023] \\
Gemini-2.5-Pro  & 20.09 [14.75, 25.43] & 3.61 [3.11, 4.10] & 193.5 [168.2, 218.7] & 0.188 [0.157, 0.219] \\
Grok-4          & 22.83 [17.68, 27.98] & 9.90 [7.68, 12.11]& 173.2 [148.0, 198.3] & 0.048 [0.030, 0.067] \\
Grok-4.1        & 24.66 [19.78, 29.54] & 9.93 [8.09, 11.78]& 196.4 [175.4, 217.3] & 0.077 [0.047, 0.107] \\
\bottomrule
\end{tabular}
\end{table}

\subsection{Grouping on Task Category Reveals Strong Model-Task Interactions}
\label{sec:task-category-results}

\begin{figure}[H]
  \centering
  \includegraphics[width=\textwidth]{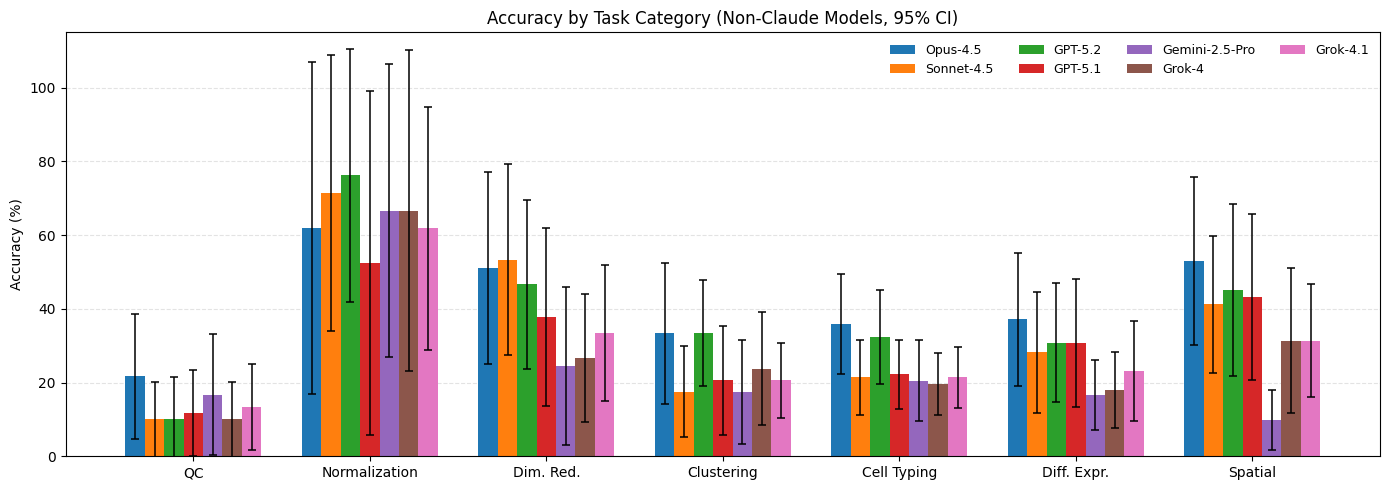}
  \caption{Model accuracy stratified by task category.}
  \label{fig:task-category}
\end{figure}

Stratifying performance by task category reveals large differences that are
not visible in aggregate accuracy (Figure~\ref{fig:task-category}, Table~\ref{tab:task-accuracy}).
Mean accuracy ranges from approximately 10–22\% on QC tasks to roughly 40–53\%
on dimensionality reduction and spatial analysis for the best-performing models.
Model rankings vary substantially across task classes: Opus-4.5 performs best
on spatial analysis (52.9\%) and ties GPT-5.2 on clustering (33.3\%), while
Sonnet-4.5 achieves the highest accuracy on dimensionality reduction (53.3\%).
In contrast, all models perform poorly on QC and cell typing, with several
models' QC confidence intervals overlapping near zero; we examine the
behavioral factors underlying these failures in the trajectory analysis below.
Gemini-2.5-Pro underperforms across all categories, most notably on spatial
analysis (9.8\%), more than 20 percentage points below the next lowest model.

\begin{table}[H]
\centering
\caption{Accuracy stratified by task category. Values are mean accuracy (\%) with 95\% confidence intervals.}
\label{tab:task-accuracy}
\resizebox{\textwidth}{!}{%
\begin{tabular}{lccccccc}
\toprule
Model & QC & Norm. & Dim.\ Red. & Clustering & Cell Typing & Diff.\ Expr. & Spatial \\
\midrule
Opus-4.5       & 21.7 [4.7,38.7]  & 61.9 [16.8,100.0] & 51.1 [25.1,77.1] & 33.3 [14.1,52.5] & 35.9 [22.2,49.6] & 37.2 [19.2,55.2] & 52.9 [30.2,75.7] \\
Sonnet-4.5     & 10.0 [0.0,20.3]  & 71.4 [34.0,100.0] & 53.3 [27.4,79.3] & 17.5 [5.1,29.8]  & 21.4 [11.3,31.4] & 28.2 [11.7,44.7] & 41.2 [22.5,59.9] \\
GPT-5.2        & 10.0 [0.0,21.4]  & 76.2 [41.9,100.0] & 46.7 [23.7,69.6] & 33.3 [18.9,47.7] & 32.5 [19.7,45.2] & 30.8 [14.7,46.9] & 45.1 [21.7,68.5] \\
GPT-5.1        & 11.7 [0.0,23.3]  & 52.4 [5.8,99.0]   & 37.8 [13.7,61.8] & 20.6 [5.9,35.4]  & 22.2 [12.8,31.6] & 30.8 [13.4,48.2] & 43.1 [20.7,65.6] \\
Gemini-2.5-Pro & 16.7 [0.3,33.1]  & 66.7 [26.9,100.0] & 24.4 [3.0,45.9]  & 17.5 [3.4,31.6]  & 20.5 [9.5,31.5]  & 16.7 [7.2,26.2]  & 9.8 [1.8,17.9]   \\
Grok-4         & 10.0 [0.0,20.3]  & 66.7 [23.1,100.0] & 26.7 [9.3,44.0]  & 23.8 [8.5,39.1]  & 19.7 [11.2,28.1] & 18.0 [7.7,28.2]  & 31.4 [11.8,51.0] \\
Grok-4.1       & 13.3 [1.6,25.1]  & 61.9 [29.0,94.9]  & 33.3 [14.9,51.8] & 20.6 [10.5,30.8] & 21.4 [13.0,29.8] & 23.1 [9.5,36.7]  & 31.4 [16.0,46.8] \\
\bottomrule
\end{tabular}%
}
\end{table}

\subsection{Grouping on Vendor Suggests Spatial Technology Dependent Performance}
\label{sec:platform-results}

\begin{figure}[H]
  \centering
  \includegraphics[width=\textwidth]{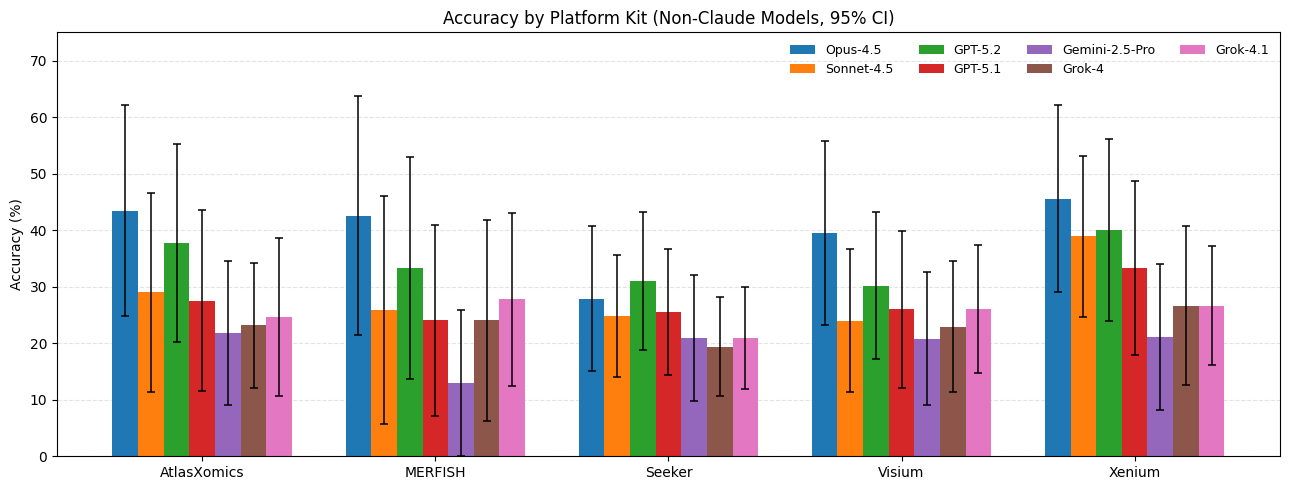}
  \caption{Model accuracy stratified by experimental platform.}
  \label{fig:platform}
\end{figure}

Accuracy varies substantially by experimental platform (Figure~\ref{fig:platform},
Table~\ref{tab:platform-accuracy}), with individual models
differing by 15–20 percentage points across datasets. Despite this variation,
relative performance is stable: Opus-4.5 leads on four of five platforms, while
GPT-5.2 leads on Seeker and ranks second elsewhere. Gemini-2.5-Pro and Grok
variants underperform across all platforms. Seeker exhibits uniformly low
accuracy across models (approximately 19–31\%), despite having the largest
number of evaluations, indicating higher intrinsic task difficulty.

\begin{table}[H]
\centering
\caption{Accuracy stratified by experimental platform. Values are mean accuracy (\%) with 95\% confidence intervals.}
\label{tab:platform-accuracy}
\resizebox{\textwidth}{!}{%
\begin{tabular}{lccccc}
\toprule
Model & AtlasXomics & MERFISH & Seeker & Visium & Xenium \\
\midrule
Opus-4.5       & 43.5 [24.8,62.1] & 42.6 [21.5,63.7] & 27.9 [15.1,40.8] & 39.6 [23.3,55.9] & 45.6 [29.1,62.1] \\
Sonnet-4.5     & 29.0 [11.4,46.5] & 25.9 [5.8,46.1]  & 24.8 [14.1,35.6] & 24.0 [11.3,36.6] & 38.9 [24.6,53.2] \\
GPT-5.2        & 37.7 [20.1,55.2] & 33.3 [13.6,53.0] & 31.0 [18.9,43.1] & 30.2 [17.1,43.3] & 40.0 [23.9,56.2] \\
GPT-5.1        & 27.5 [11.5,43.6] & 24.1 [7.2,41.0]  & 25.6 [14.4,36.7] & 26.0 [12.1,39.9] & 33.3 [18.0,48.7] \\
Gemini-2.5-Pro & 21.7 [9.0,34.5]  & 13.0 [0.1,25.9]  & 20.9 [9.7,32.1]  & 20.8 [9.1,32.6]  & 21.1 [8.3,34.0]  \\
Grok-4         & 23.2 [12.2,34.2] & 24.1 [6.3,41.9]  & 19.4 [10.6,28.1] & 22.9 [11.3,34.5] & 26.7 [12.7,40.7] \\
Grok-4.1       & 24.6 [10.8,38.5] & 27.8 [12.5,43.1] & 20.9 [12.0,29.9] & 26.0 [14.7,37.4] & 26.7 [16.1,37.2] \\
\bottomrule
\end{tabular}%
}
\end{table}

\subsection{Agent Harness Choice Materially Affects Benchmark Performance}
\label{sec:harness-comparison}

\begin{figure}[H]
  \centering
  \includegraphics[width=0.8\textwidth]{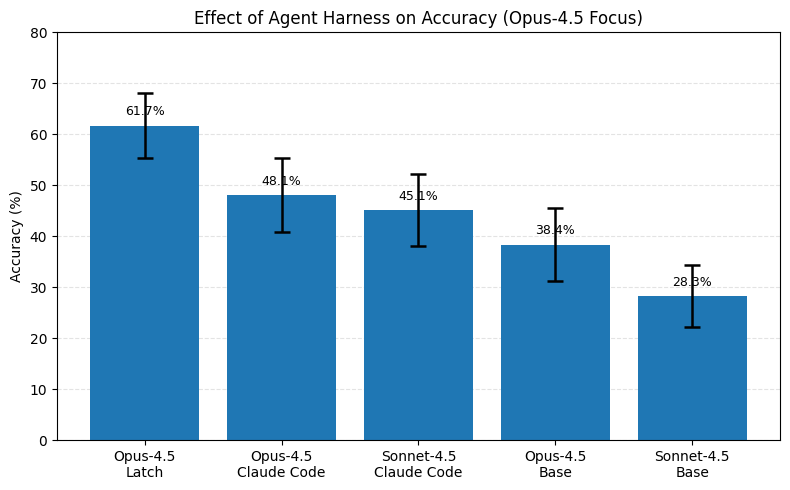}
  \caption{Aggregate accuracy comparison across agent harnesses.}
  \label{fig:harness-accuracy}
\end{figure}

We examined the effect of the agent \emph{harness}---the execution environment,
tools, prompts, and programmatic control flow that wrap a base model---on
performance (Figure~\ref{fig:harness-accuracy}, Table~\ref{tab:harness-aggregate-accuracy}).
Comparing Opus-4.5 across a slightly modified Mini-SWE-Bench setup, Claude Code,
and the Latch agent shows that harness design has a large effect on outcomes.
Within the Latch harness, Opus-4.5 attains 61.7\% overall accuracy, exceeding
Opus-4.5 with Claude Code (48.1\%) and the base configuration (38.4\%). The
absolute uplift from base to Latch is 23.3 percentage points, and from Claude
Code to Latch is 13.6 points, exceeding the Opus--Sonnet gap under Claude Code
(48.1\% vs.\ 45.1\%).

\begin{figure}[H]
  \centering
  \includegraphics[width=\textwidth]{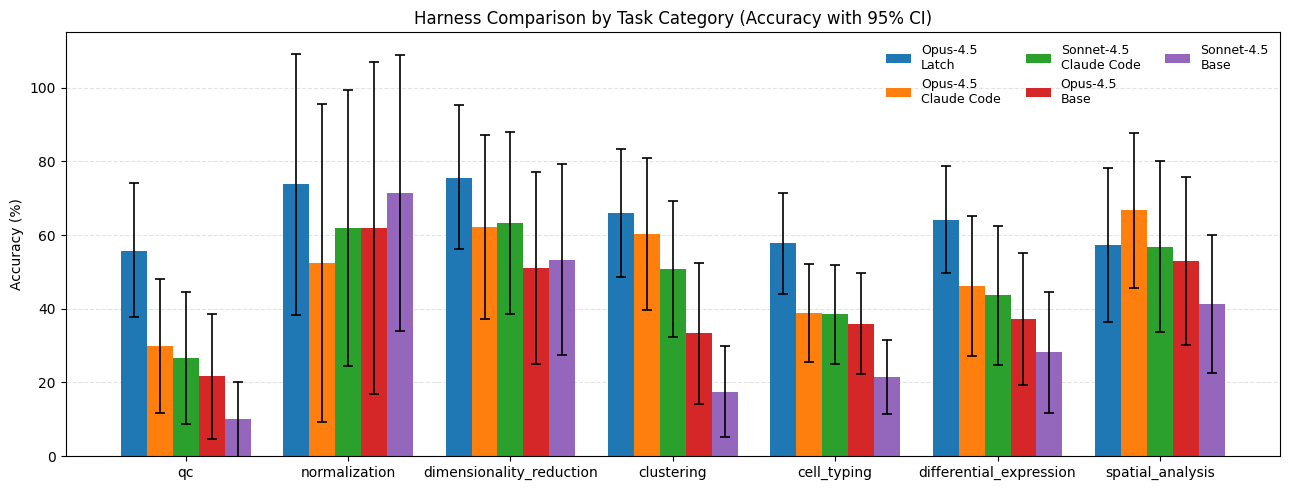}
  \caption{Harness performance stratified by task category.}
  \label{fig:harness-by-task}
\end{figure}

Task-stratified results (Figure~\ref{fig:harness-by-task}, Table~\ref{tab:harness-task-accuracy})
suggest that the Latch and Claude Code harnesses primarily improve performance
on tasks that require many steps and intermediate programming. Clustering
(65.9\% vs.\ 33.3\%), differential expression (64.1\% vs.\ 37.2\%), and
dimensionality reduction (75.6\% vs.\ 51.1\%) are representative examples
(Latch vs.\ Opus-4.5 base).

\begin{figure}[H]
  \centering
  \includegraphics[width=\textwidth]{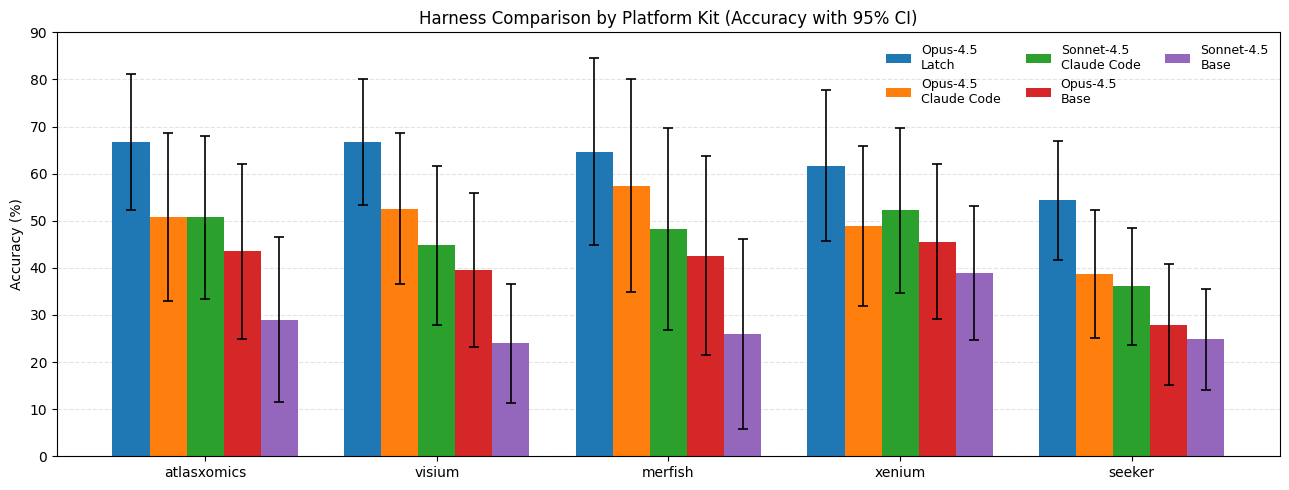}
  \caption{Harness performance stratified by experimental platform.}
  \label{fig:harness-by-kit}
\end{figure}

Harness improvements are also robust across spatial platforms
(Figure~\ref{fig:harness-by-kit}, Table~\ref{tab:harness-platform-accuracy}),
mirroring patterns seen in other analyses. This suggests that effective
harnesses provide models with a general ability to manipulate and reason about
spatial data, likely by stabilizing exploration and analysis over longer time
horizons. Platform-specific difficulty, however, will require special care
(e.g., deliberate prompt engineering and tool selection) and is unlikely to
resolve accidentally with base model improvements alone.

\begin{table}[H]
\centering
\caption{Aggregate accuracy across agent harnesses.
Reported values are mean accuracy with 95\% confidence intervals.
Latch results are available only for Opus-4.5.}
\label{tab:harness-aggregate-accuracy}
\begin{tabular}{lcc}
\toprule
Model / Harness & Accuracy (\%) & 95\% CI \\
\midrule
Opus-4.5 (Latch)        & 61.7 & [55.3, 68.1] \\
Opus-4.5 (Claude Code)  & 48.1 & [40.9, 55.3] \\
Sonnet-4.5 (Claude Code)& 45.1 & [38.0, 52.2] \\
Opus-4.5 (Base)         & 38.4 & [31.3, 45.4] \\
Sonnet-4.5 (Base)       & 28.3 & [22.2, 34.4] \\
\bottomrule
\end{tabular}
\end{table}

\begin{table}[H]
\centering
\caption{Accuracy stratified by task category for different agent harnesses.
Values are mean accuracy (\%) with 95\% confidence intervals.
Latch results are available only for Opus-4.5.}
\label{tab:harness-task-accuracy}
\resizebox{\textwidth}{!}{%
\begin{tabular}{lccccccc}
\toprule
Condition & QC & Norm. & Dim.\ Red. & Clustering & Cell Typing & Diff.\ Expr. & Spatial \\
\midrule
Opus-4.5 (Latch) & 55.8 [37.7, 74.0] & 73.8 [38.4, 100.0] & 75.6 [56.1, 95.2] & 65.9 [48.5, 83.3] & 57.7 [44.0, 71.4] & 64.1 [49.6, 78.6] & 57.4 [36.5, 78.3] \\
Opus-4.5 (Claude Code) & 30.0 [11.8, 48.2] & 52.4 [9.3, 95.5] & 62.2 [37.2, 87.3] & 60.3 [39.6, 81.0] & 38.9 [25.6, 52.1] & 46.2 [27.1, 65.2] & 66.7 [45.7, 87.7] \\
Sonnet-4.5 (Claude Code) & 26.7 [8.7, 44.6] & 61.9 [24.5, 99.4] & 63.3 [38.6, 88.1] & 50.8 [32.5, 69.1] & 38.5 [25.0, 51.9] & 43.6 [24.6, 62.5] & 56.9 [33.6, 80.2] \\
Opus-4.5 (Base) & 21.7 [4.7, 38.7] & 61.9 [16.8, 100.0] & 51.1 [25.1, 77.1] & 33.3 [14.1, 52.5] & 35.9 [22.2, 49.6] & 37.2 [19.2, 55.2] & 52.9 [30.2, 75.7] \\
Sonnet-4.5 (Base) & 10.0 [0.0, 20.3] & 71.4 [34.0, 100.0] & 53.3 [27.4, 79.3] & 17.5 [5.1, 29.8] & 21.4 [11.3, 31.4] & 28.2 [11.7, 44.7] & 41.2 [22.5, 59.9] \\
\bottomrule
\end{tabular}%
}
\end{table}

\begin{table}[H]
\centering
\caption{Accuracy stratified by spatial transcriptomics platform.
Values are mean accuracy (\%) with 95\% confidence intervals.
Latch results are available only for Opus-4.5.}
\label{tab:harness-platform-accuracy}
\resizebox{\textwidth}{!}{%
\begin{tabular}{lccccc}
\toprule
Condition & AtlasXomics & Visium & MERFISH & Xenium & Seeker \\
\midrule
Opus-4.5 (Latch)       & 66.7 [52.3, 81.1] & 66.7 [53.4, 80.0] & 64.7 [44.9, 84.5] & 61.7 [45.6, 77.7] & 54.3 [41.6, 66.9] \\
Opus-4.5 (Claude Code) & 50.7 [32.9, 68.6] & 52.6 [36.5, 68.7] & 57.4 [34.8, 80.0] & 48.9 [32.0, 65.8] & 38.8 [25.2, 52.4] \\
Sonnet-4.5 (Claude Code) & 50.7 [33.4, 68.0] & 44.8 [27.9, 61.7] & 48.2 [26.7, 69.6] & 52.2 [34.7, 69.7] & 36.1 [23.7, 48.4] \\
Opus-4.5 (Base)        & 43.5 [24.8, 62.1] & 39.6 [23.3, 55.9] & 42.6 [21.5, 63.7] & 45.6 [29.1, 62.1] & 27.9 [15.1, 40.8] \\
Sonnet-4.5 (Base)      & 29.0 [11.4, 46.5] & 24.0 [11.3, 36.6] & 25.9 [5.8, 46.1]  & 38.9 [24.6, 53.2] & 24.8 [14.1, 35.6] \\
\bottomrule
\end{tabular}%
}
\end{table}

\subsection{Agent Trajectories Reveal Distinct Behavioral Patterns}
\label{sec:agent-behavior}

Manual inspection of agent trajectories (session logs containing reasoning
traces, tool invocations, and terminal standard output and error) revealed
qualitative behavioral patterns that explain performance differences between
frontier models more mechanistically (Table~\ref{tab:agent-behavior-summary}).

\begin{table}[H]
\centering
\caption{Summary of agent behavioral characteristics across models.
Metrics characterize typical control flow, error patterns, and domain calibration
observed across SpatialBench evaluations.}
\label{tab:agent-behavior-summary}
\begin{tabular}{lcccccc}
\toprule
Model & Accuracy (\%) & Mean Steps & Max Steps & 6+ Steps (\%) & Format Errors & min\_genes (med.) \\
\midrule
Opus-4.5   & 38.4 & 2.8 & 13  & 10.6 & 0.04 & 10  \\
GPT-5.2    & 34.0 & 2.1 & 9   & 2.7  & 0.00 & 200 \\
Sonnet-4.5 & 28.3 & 2.4 & 17  & 7.8  & 0.05 & 200 \\
GPT-5.1    & 27.4 & 2.4 & 7   & 4.6  & 0.01 & 200 \\
Grok-4.1   & 24.7 & 9.9 & 100 & 33.1 & 7.16 & 200 \\
Grok-4     & 22.8 & 9.8 & 100 & 23.9 & 6.74 & 200 \\
Gemini-2.5 & 20.1 & 3.6 & 26  & 21.5 & 0.06 & 100 \\
\bottomrule
\end{tabular}
\end{table}

\begin{figure}[H]
  \centering
  \includegraphics[width=\textwidth]{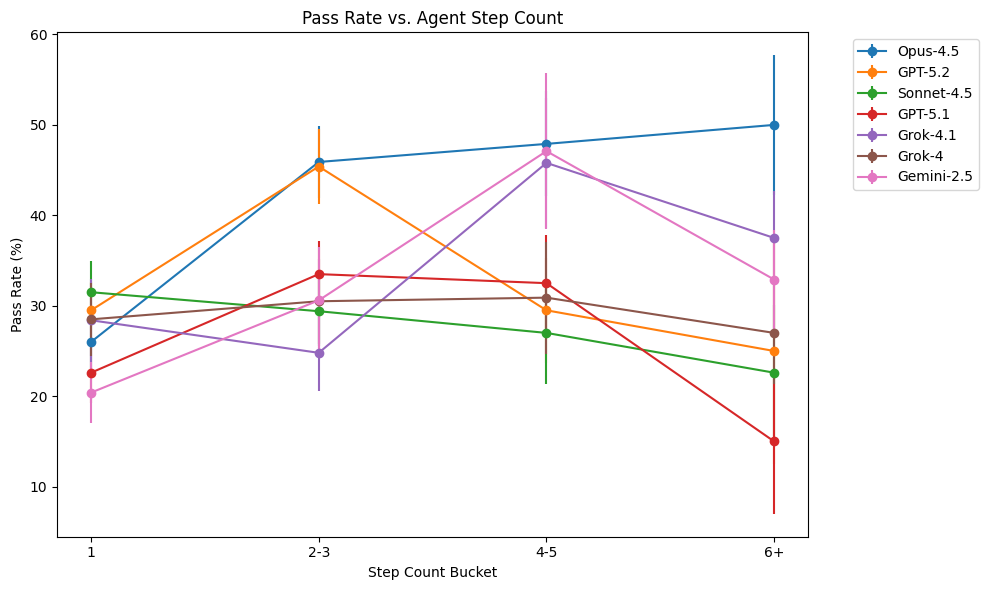}
  \caption{Pass rate stratified by agent step count.}
  \label{fig:passrate-by-steps}
\end{figure}

\paragraph{Steps.}
Claude and GPT models complete tasks in approximately 2–3 steps on average
(mean 2.1–2.8), whereas Grok variants require nearly four times as many steps
(mean 9.8–9.9), with Gemini intermediate (3.6). Extremely high step counts
generally indicate thrashing: all 119 instances of 100-step limit exhaustion
occurred exclusively in Grok runs, and all resulted in failure.
More steps are not always bad, so we also examine step productivity as a more
informative metric (Figure~\ref{fig:passrate-by-steps}, Table~\ref{tab:passrate-by-steps}).
For Opus-4.5, pass rate increases monotonically with step
count, rising from 26.0\% for one-step runs to 50.0\% for runs with six or more
steps, indicating capacity for productive exploration. In contrast, Grok-4
exhibits nearly flat pass rates across step buckets ($\sim$27--31\%),
suggesting retry-driven behavior rather than incremental progress; Grok-4.1
shows some uplift at intermediate step counts but still suffers from high
step exhaustion and format error rates. Gemini
exhibits bimodal behavior: while 50.6\% of runs complete in a single step,
21.5\% require six or more steps, reflecting inconsistent problem-solving
strategies or prolonged failure loops.

\begin{table}[H]
\centering
\caption{Pass rate stratified by agent step count.
Values are mean pass rate (\%) with standard error (SE) in parentheses; $n$ denotes the number of evaluation instances in each bucket.}
\label{tab:passrate-by-steps}
\resizebox{\textwidth}{!}{%
\begin{tabular}{lcccc}
\toprule
Model & 1 step & 2--3 steps & 4--5 steps & 6+ steps \\
\midrule
Opus-4.5    & 26.0 (3.5) [$n$=154] & 45.9 (4.0) [$n$=157] & 47.9 (5.8) [$n$=73] & 50.0 (7.7) [$n$=42] \\
GPT-5.2     & 29.5 (3.1) [$n$=217] & 45.4 (4.2) [$n$=141] & 29.5 (5.8) [$n$=61] & 25.0 (12.5) [$n$=12] \\
Sonnet-4.5  & 31.5 (3.4) [$n$=184] & 29.4 (3.8) [$n$=143] & 27.0 (5.6) [$n$=63] & 22.6 (7.5) [$n$=31] \\
GPT-5.1     & 22.6 (3.3) [$n$=164] & 33.5 (3.7) [$n$=164] & 32.5 (5.3) [$n$=77] & 15.0 (8.0) [$n$=20] \\
Grok-4.1    & 28.4 (4.6) [$n$=95]  & 24.8 (4.2) [$n$=105] & 45.8 (7.2) [$n$=48] & 37.5 (5.2) [$n$=88] \\
Grok-4      & 28.5 (4.0) [$n$=130] & 30.5 (4.7) [$n$=95]  & 30.9 (6.2) [$n$=55] & 27.0 (5.6) [$n$=63] \\
Gemini-2.5  & 20.4 (3.4) [$n$=142] & 30.6 (5.9) [$n$=62]  & 47.1 (8.6) [$n$=34] & 32.9 (5.5) [$n$=73] \\
\bottomrule
\end{tabular}%
}
\end{table}

\paragraph{Instruction following.}
Instruction-following failures strongly differentiate model families. Grok
variants average more than seven format errors per evaluation, accumulating
over 6,000 total errors across both variants. These errors consume steps on
syntax compliance rather than analysis and frequently lead to runaway retry
loops. In contrast, GPT-5.2 produces zero format errors, and Claude and Gemini
models exhibit near-perfect compliance ($\leq$0.06 errors per evaluation). Manual
quality control confirmed that these formatting issues reflect legitimate
reasoning failures rather than ambiguities in harness prompts or
infrastructure bugs.

\paragraph{Domain knowledge.}
Models exhibit systematic differences in domain calibration, particularly for
quality control thresholds. In some targeted and imaging-based spatial assays—where per-cell gene counts are intrinsically low—reasonable \texttt{min\_genes} thresholds lie between 5 and 20. Opus-4.5 applies spatially calibrated cutoffs (median 10), whereas other models default to scRNA-seq–like thresholds (median 100–200). This
calibration difference aligns with Opus-4.5's substantially higher QC pass
rate ($\sim$25\%) relative to GPT and Grok models ($\sim$5\%).

\paragraph{Productive exploration.}
Models differ in how effectively they explore and use intermediate findings.
While Opus-4.5 checks \texttt{adata.uns} most frequently (34.0\% of
evaluations vs.\ 17–18\% for GPT models), the key distinction is utilization:
when Opus inspects \texttt{adata.uns}, its pass rate increases by 26
percentage points (56.6\% vs.\ 30.7\%), compared to just 4–6 percentage
points for Grok despite similar inspection rates ($\sim$26\%). This suggests that
finding pre-computed results is insufficient—models must also recognize their
relevance and integrate them into downstream analysis.

\paragraph{Behavioral regimes by model family.}
Taken together, these patterns reveal distinct behavioral regimes. Opus-4.5
benefits substantially from multi-step execution (pass rate increases from
26\% at one step to 50\% at six or more steps) and productively applies domain
knowledge such as spatially appropriate thresholds. GPT models (5.1, 5.2)
execute efficiently with near-perfect instruction following but leverage less
domain-specific knowledge and productive exploration. Grok models (4, 4.1)
struggle with instruction compliance, leading to high step counts without
corresponding accuracy gains. Gemini-2.5-Pro displays inconsistent behavior:
despite moderate step counts and low format error rates, it achieves the
lowest overall accuracy, suggesting fundamental gaps in domain knowledge and
analytical reasoning.

\section{Methods}
\label{sec:methods}

\subsection{Overview of Problem Construction}
\label{sec:problem-construction}

SpatialBench is constructed from real spatial transcriptomics analysis workflows~\citep{moses2022museum}
provided by domain scientists and technology vendors. For each workflow, we
identify analysis steps that (i) arise in practice, (ii) require interaction with
messy, high-dimensional data objects, and (iii) yield a biologically meaningful
intermediate result that can be verified automatically. We then convert each
step into an evaluation problem by snapshotting the analysis state immediately
prior to the target step and writing a natural-language prompt that asks for a
scientific conclusion or decision that a competent analyst would make from that
state (e.g., choosing a QC threshold, identifying a cell type from marker
patterns, or determining which populations separate along an embedding axis).

Each candidate problem is iteratively refined with the contributing experts to
ensure that the intended answer is stable under reasonable analysis choices and
does not depend on a single implementation detail. Finally, we apply manual
quality control that includes inspection of multi-run agent trajectories and
adversarial “shortcut” attempts (e.g., agents instructed to guess from prior
knowledge without interacting with the data) to remove problems that can be
solved without performing the intended analysis.

\subsection{Anatomy of a Problem}
\label{sec:problem-anatomy}

Each SpatialBench item consists of (1) a workspace containing the relevant data
objects (typically an \texttt{AnnData} object~\citep{virshup2024anndata,wolf2018scanpy,palla2022squidpy,marconato2024spatialdata} and associated files), (2) a task
prompt describing the goal at a scientific level, (3) an answer format
specification (to enforce a gradeable output), and (4) a deterministic grader
that maps the agent’s final answer to pass/fail (and, when useful, auxiliary
diagnostics). Importantly, the input state is a \emph{snapshot}: it reflects the
information that would be available to an analyst at that point in an
end-to-end workflow (e.g., prior preprocessing already applied, but the target
analysis not yet performed).

To encourage empirical interaction with the dataset, prompts are written so that
domain knowledge alone is insufficient: correct answers typically require
computing summary statistics, inspecting distributions, testing marker genes,
or reading structured results stored in the analysis object (e.g.,
\texttt{adata.obs}, \texttt{adata.var}, \texttt{adata.uns}). This design reduces
the viability of pattern matching and promotes behaviors closer to practical
analysis.

\subsection{Properties of a Good SpatialBench Test}
\label{sec:test-properties}

We follow three core design criteria. \textbf{Verifiability:} each task must
admit an automatically checkable success condition (e.g., a set overlap, a
numerical interval check, or a structured label match) implemented as a
deterministic grader. \textbf{Scientific durability:} the intended conclusion
should remain correct across reasonable algorithmic choices and hyperparameters,
so the benchmark measures biological interpretation rather than brittle
implementation details. \textbf{Anti-shortcut structure:} tasks are phrased to
require interacting with the provided data artifacts, and we remove items that
can be reliably solved via trivial heuristics or prior knowledge.

These criteria lead to problems that are intentionally closer to real analysis:
agents must navigate data objects, generate plots or summaries, and make
context-aware decisions under noise and ambiguity, while still producing outputs
that can be graded reproducibly.

\subsection{Verifiable Graders}
\label{sec:graders}

Each problem is paired with a grader that evaluates whether the agent recovered
the key biological result in a verifiable manner. Graders operate on the
agent’s final structured answer and do not inspect intermediate reasoning. We
use several grader families, including: exact or near-exact match for discrete
labels, set similarity (e.g., Jaccard) for gene lists or cluster membership,
interval-based checks for numeric thresholds, and consistency checks for
multi-field outputs. Where appropriate, graders include conservative tolerances
to avoid over-penalizing scientifically equivalent answers (e.g., synonyms for
cell types, or small deviations around an empirically chosen threshold).

To reduce brittleness, graders are validated during problem development by
running multiple independent baselines and manually inspecting failures to
distinguish grader bugs from legitimate reasoning errors. Problems whose grading
logic cannot be made robust without re-introducing subjectivity are excluded
from the benchmark.

\subsection{Agent Harnesses and Execution Environment}
\label{sec:harnesses}

We evaluate models under multiple \emph{harnesses}~\citep{yang2024sweagent,yao2023react,liu2023agentbench,zhou2023webarena}, where a harness denotes the
full execution wrapper around a base model: the system prompt, available tools,
control flow (planning and retry policies), answer schema enforcement, and the
runtime environment used to execute code. All harnesses provide an interactive
compute setting with access to common scientific Python tooling and the local
workspace containing the problem’s data snapshot. Harnesses differ primarily in
their prompting strategy, tool routing, and how they structure multi-step work
(e.g., whether they enforce intermediate checks, how they respond to errors, and
how they decide when to stop).

To make comparisons meaningful, we use a consistent evaluation budget per item
(step limits and timeouts) within each harness condition and record detailed
session logs (tool calls, stdout/stderr, intermediate artifacts) to support
trajectory-level analysis of failure modes.

\subsection{Data Infrastructure: Running SpatialBench at Scale}
\label{sec:data-infrastructure}

SpatialBench is executed as batched evaluations in isolated workspaces to ensure
reproducibility. Each evaluation materializes a clean workspace containing the
problem snapshot and executes the agent harness under fixed resource limits.
Outputs include the agent’s final structured answer, the grader decision, and a
complete trajectory log. All runs are keyed by model version, harness version,
and benchmark revision to support exact reruns and ablations.

\subsection{Statistical Design}
\label{sec:statistical-design}

SpatialBench uses the evaluation item (\emph{eval}) as the statistical unit. For
each eval $i \in \{1,\dots,n\}$ we run the agent $K=3$ times under the same model
and harness condition. Each run receives a binary outcome from the deterministic
grader, $s_{i,r} \in \{0,1\}$ for run $r \in \{1,\dots,K\}$.

\paragraph{Terminology.}
We use \emph{pass rate} to refer to the fraction of runs that pass the grader,
and \emph{accuracy} for the aggregate estimate $\hat{\mu}$ over eval-level means.
A \emph{step} is a single agent action (one model invocation plus any resulting
tool calls); step counts measure agent effort per evaluation.

\paragraph{Stage 1: per-eval mean score.}
We first compute the per-eval mean pass rate
\[
\bar{s}_i = \frac{1}{K}\sum_{r=1}^{K} s_{i,r},
\]
so each eval contributes a value in $[0,1]$ (e.g., $\bar{s}_i = 2/3$ if two of
three runs pass).

\paragraph{Stage 2: confidence intervals over eval means.}
We treat the $\{\bar{s}_i\}_{i=1}^{n}$ as independent observations and compute
the aggregate accuracy
\[
\hat{\mu} = \frac{1}{n}\sum_{i=1}^{n} \bar{s}_i.
\]
Let
\[
\hat{\sigma}^2 = \frac{1}{n-1}\sum_{i=1}^{n}(\bar{s}_i - \hat{\mu})^2
\quad\text{and}\quad
\mathrm{SE}(\hat{\mu}) = \sqrt{\frac{\hat{\sigma}^2}{n}}.
\]
We report 95\% confidence intervals as
\[
\hat{\mu} \pm t_{0.975,\,n-1}\,\mathrm{SE}(\hat{\mu}),
\]
where $t_{0.975,\,n-1}$ is the 97.5th percentile of a Student-$t$ distribution
with $n-1$ degrees of freedom.

\paragraph{Stratified results.}
For breakdowns by task category or platform, we apply the same two-stage
procedure to the subset of evals in the stratum (recomputing $n$, $\hat{\mu}$,
$\hat{\sigma}^2$, and the corresponding $t$ critical value).

\paragraph{Efficiency metrics.}
Steps, latency, and cost are summarized analogously by first averaging the metric
within each eval across the $K$ runs, then computing a $t$-based confidence
interval across eval-level means.

\section{Discussion}
\label{sec:discussion}

SpatialBench measures a capability between “knowing biology” and “writing
code”: extracting biological insight from messy, real-world spatial datasets.
These tasks are scientifically central and represent a distinct challenge for
frontier agentic systems, requiring a blend of programming, data analysis, and
domain reasoning. Across 146 verifiable problems, today’s frontier models
remain unreliable, leaving substantial room for progress.

The path forward appears to be a long tail of tractable engineering and
calibration. Stratifying results by task category and platform, and inspecting
agent trajectories, reveals concrete mechanisms underpinning failures and
successes. Some systems fail primarily through instruction-following and
output-format errors, wasting steps on compliance rather than analysis. More
broadly, the uniformly poor performance on quality control and cell typing
indicates that steps requiring contextual—often tacit—scientific judgment are
those current models handle least well. Transferring general-purpose coding or
data-analysis skills is not sufficient; models will likely need exposure to
representative spatial workflows across tissues, diseases, and assay types,
either in training data or via harness-level scaffolding (tools, retrieval, and
calibration routines).

Harness design emerges as a major determinant of performance. The same base
model exhibits large accuracy variation across harnesses, indicating that what
is often treated as “glue code”—tools, prompts, control flow, execution
infrastructure, and verification—can unlock or suppress capability. Progress in
practical biological agents will therefore require joint optimization of model
and harness. Benchmarks should report harness details as rigorously as model
versions, and evaluations should treat the full agent stack as the unit of
study.

Performance also depends strongly on spatial technology. We observe 15–20 point
swings across platforms for the same model, reflecting that spatial biology is
a collection of diverse assays with distinct artifacts and conventions.
Reliable agents will likely require platform-aware context, assay-specific
tools, and self-calibration heuristics rather than a one-size-fits-all
workflow.

SpatialBench has limitations. Deterministic graders enable verifiable
evaluation, but can be brittle, and important nuance is sometimes lost when
scientific judgment is discretized into automatically checkable outputs. In
addition, each problem is a snapshotted step of an end-to-end workflow;
longer-horizon analyses involve compounding errors and iterative revision
(e.g., revisiting QC thresholds after poor clustering), which are not yet fully
captured here.

We hope SpatialBench serves both as a measurement tool and a diagnostic lens,
enabling the community to develop agents that interact with real spatial
datasets faithfully, transparently, and reproducibly. SpatialBench is intended
as a first focused contribution toward a broader benchmark family spanning
major biological data classes. More broadly, we view benchmarks as evolving
specifications of computational biology workflows—supporting test-driven
development of agent systems whose behavior can be improved through both model
training and harness engineering.

\section*{Data and Code Availability}
\label{sec:availability}

All benchmark materials are publicly available at \url{https://github.com/latchbio/spatialbench}:

\begin{itemize}
    \item \texttt{results/} --- Aggregate benchmark results with 95\% confidence intervals (overall, per-task, per-platform)
    \item \texttt{evals\_canonical/} --- 10 canonical evaluation examples spanning all task categories
    \item \texttt{spatialbench/graders/} --- Grader implementations for all five grader families
\end{itemize}

\section*{Author Contributions}

K.W. conceived the project, designed the benchmark methodology, built the
evaluation infrastructure, constructed evaluations, and wrote the manuscript.
Z.Y., H.M., and H.L. constructed evaluations and contributed to benchmark
validation.

\bibliographystyle{unsrtnat}
\bibliography{references}

\appendix
\section*{A.1 Full Benchmark Inventory}

This section provides the complete inventory of all 146 evaluations in SpatialBench, organized by platform and task category.

\subsection*{Summary Statistics}

SpatialBench comprises 146 evaluations spanning 5 spatial transcriptomics platforms and 7 task categories. Table~\ref{tab:benchmark-counts} shows the distribution of evaluations across platforms and task categories.

\begin{table}[h]
\centering
\caption{Distribution of evaluations across platforms and task categories.}
\label{tab:benchmark-counts}
\small
\begin{tabular}{lcccccccl}
\toprule
Platform & QC & Norm. & Dim. Red. & Clust. & Cell Type & Diff. Expr. & Spatial & \textbf{Total} \\
\midrule
AtlasXomics & 6 & — & — & 1 & 11 & 5 & — & \textbf{23} \\
MERFISH & 2 & 2 & 2 & 3 & 4 & 4 & 2 & \textbf{19} \\
Xenium & 3 & 2 & 3 & 4 & 6 & 7 & 5 & \textbf{30} \\
Visium & 5 & 1 & 4 & 5 & 8 & 4 & 5 & \textbf{32} \\
Seeker & 4 & 2 & 6 & 9 & 10 & 6 & 6 & \textbf{43} \\
\midrule
\textbf{Total} & 20 & 7 & 15 & 22 & 39 & 26 & 18 & \textbf{146} \\
\bottomrule
\end{tabular}
\end{table}

\subsubsection*{Grader Distribution}

Evaluations use five grader families to assess agent outputs:

\begin{itemize}
    \item \textbf{MultipleChoice} (45 evals): Single correct answer from predefined options
    \item \textbf{MarkerGenePrecisionRecall} (45 evals): Precision@K for gene list recovery
    \item \textbf{NumericTolerance} (36 evals): Numeric values within absolute tolerance bounds
    \item \textbf{LabelSetJaccard} (18 evals): Set overlap using Jaccard similarity threshold
    \item \textbf{DistributionComparison} (3 evals): Cosine similarity for distribution matching
\end{itemize}

\subsubsection*{Tissue Coverage}

The benchmark covers five tissue types across the platforms:
\begin{itemize}
    \item \textbf{Kidney} (Xenium): 27 evaluations — acute kidney injury, fibrosis, tubular cell states
    \item \textbf{Ovary} (Seeker): 43 evaluations — follicle development, granulosa cell subtypes, ovulation response
    \item \textbf{Bone} (Visium): 32 evaluations — bone remodeling, trabecular microenvironment, immune infiltration
    \item \textbf{Brain} (MERFISH, AtlasXomics): 30 evaluations — aging signatures, glial cell states, neuronal populations
    \item \textbf{Liver} (AtlasXomics): 15 evaluations — cirrhosis, hepatic stellate cells, fibrosis pathways
\end{itemize}

\subsection*{Complete Evaluation Inventory}

Table~\ref{tab:full-inventory} provides the complete list of all 146 evaluations with their metadata. Description provides a concise summary of each evaluation task. Platform indicates the spatial technology. Task denotes the analysis category. Tissue specifies the biological sample type. Grader indicates the evaluation method: MCQ (multiple choice), P@K (precision at K for gene lists), Numeric (numeric tolerance), Jaccard (label set Jaccard similarity), or Cosine (distribution cosine similarity). Params shows key grader parameters where applicable.

\begin{longtable}{p{3.5cm}lllll}
\caption{Complete inventory of SpatialBench evaluations.}
\label{tab:full-inventory} \\
\toprule
Description & Platform & Task & Tissue & Grader & Params \\
\midrule
\endfirsthead
\multicolumn{6}{c}{\tablename\ \thetable{} -- continued} \\
\toprule
Description & Platform & Task & Tissue & Grader & Params \\
\midrule
\endhead
\midrule
\multicolumn{6}{r}{Continued on next page} \\
\endfoot
\bottomrule
\endlastfoot
Astrocyte clusters & AtlasXomics & Cell Type & Brain & Jaccard & J$\geq$0.5 \\
Astrocyte markers & AtlasXomics & Cell Type & Brain & P@K & P@5 \\
Cell type distribution & AtlasXomics & Cell Type & Brain & Numeric & $\pm$10.0 \\
Endothelial clusters & AtlasXomics & Cell Type & Brain & Jaccard & J$\geq$0.5 \\
Endothelial markers & AtlasXomics & Cell Type & Brain & P@K & P@5 \\
Microglia markers & AtlasXomics & Cell Type & Brain & P@K & P@5 \\
Neuron clusters & AtlasXomics & Cell Type & Brain & Jaccard & J$\geq$0.5 \\
Neuron markers & AtlasXomics & Cell Type & Brain & P@K & P@5 \\
Oligodendrocyte clusters & AtlasXomics & Cell Type & Brain & Jaccard & J$\geq$0.5 \\
Oligodendrocyte markers & AtlasXomics & Cell Type & Brain & P@K & P@5 \\
Stellate cell clusters & AtlasXomics & Cell Type & Liver & Jaccard & J$\geq$0.5 \\
Fibrosis clustering & AtlasXomics & Clust. & Liver & P@K & P@5 \\
Confound detection & AtlasXomics & Diff. Expr. & — & MCQ & — \\
Disease signature & AtlasXomics & Diff. Expr. & Liver & P@K & P@5 \\
Hepatocyte DE & AtlasXomics & Diff. Expr. & Liver & P@K & P@5 \\
Lost pathway analysis & AtlasXomics & Diff. Expr. & Liver & MCQ & — \\
Pathway analysis & AtlasXomics & Diff. Expr. & Liver & MCQ & — \\
Cell filtering & AtlasXomics & QC & Liver & Numeric & $\pm$3892 \\
FRiP filtering & AtlasXomics & QC & Liver & Numeric & $\pm$6265 \\
Fragment count filtering & AtlasXomics & QC & Liver & Numeric & $\pm$1960 \\
Fragment distribution & AtlasXomics & QC & Liver & Numeric & $\pm$8.0 \\
Mitochondrial filtering & AtlasXomics & QC & Liver & Numeric & $\pm$1709 \\
TSS enrichment filtering & AtlasXomics & QC & Liver & Numeric & $\pm$2741 \\
\midrule
Astrocyte markers & MERFISH & Cell Type & Brain & P@K & P@5 \\
Cell type distribution & MERFISH & Cell Type & Brain & Numeric & $\pm$5.0 \\
Microglia markers & MERFISH & Cell Type & Brain & P@K & P@5 \\
Spatial heterogeneity & MERFISH & Cell Type & Brain & MCQ & — \\
Astrocyte clustering & MERFISH & Clust. & Brain & MCQ & — \\
Cluster enrichment & MERFISH & Clust. & Brain & Numeric & $\pm$0.1 \\
Niche cluster markers & MERFISH & Clust. & Brain & P@K & P@5 \\
Aging/inflammatory DE & MERFISH & Diff. Expr. & Brain & MCQ & — \\
Aging/inflammatory DE & MERFISH & Diff. Expr. & Brain & MCQ & — \\
Aging/inflammatory DE & MERFISH & Diff. Expr. & Brain & P@K & P@5 \\
Cell type changes & MERFISH & Diff. Expr. & Brain & Jaccard & J$\geq$0.5 \\
PCA cell separation & MERFISH & Dim. Red. & Brain & MCQ & — \\
PCA cell separation & MERFISH & Dim. Red. & Brain & MCQ & — \\
Log z-score normalization & MERFISH & Norm. & Brain & Numeric & $\pm$0.05 \\
Scaling normalization & MERFISH & Norm. & Brain & Numeric & $\pm$80 \\
Min UMI filtering & MERFISH & QC & Brain & Numeric & $\pm$50 \\
Min genes filtering & MERFISH & QC & Brain & Numeric & $\pm$50 \\
Proximity analysis & MERFISH & Spatial & Brain & MCQ & — \\
Spatial activation & MERFISH & Spatial & Brain & MCQ & — \\
\midrule
Cell subtyping & Seeker & Cell Type & Ovary & Numeric & $\pm$10.0 \\
Cell type distribution & Seeker & Cell Type & Ovary & Numeric & $\pm$15.0 \\
Endothelial clusters & Seeker & Cell Type & Ovary & Jaccard & J$\geq$0.5 \\
Endothelial markers & Seeker & Cell Type & Ovary & P@K & P@5 \\
Granulosa clusters & Seeker & Cell Type & Ovary & Jaccard & J$\geq$0.5 \\
Granulosa markers & Seeker & Cell Type & Ovary & P@K & P@5 \\
Mesenchymal clusters & Seeker & Cell Type & Ovary & Jaccard & J$\geq$0.5 \\
Mesenchymal markers & Seeker & Cell Type & Ovary & P@K & P@5 \\
Oocyte clusters & Seeker & Cell Type & Ovary & Jaccard & J$\geq$0.5 \\
Oocyte markers & Seeker & Cell Type & Ovary & P@K & P@5 \\
Antral cluster markers & Seeker & Clust. & Ovary & P@K & P@5 \\
Antral cluster markers & Seeker & Clust. & Ovary & P@K & P@5 \\
Atretic cluster markers & Seeker & Clust. & Ovary & P@K & P@5 \\
Cumulus cluster markers & Seeker & Clust. & Ovary & P@K & P@5 \\
HVG selection & Seeker & Clust. & Ovary & P@K & P@5 \\
Major subtype ID & Seeker & Clust. & Ovary & MCQ & — \\
Mitotic cluster markers & Seeker & Clust. & Ovary & P@K & P@5 \\
Spatial cluster coherence & Seeker & Clust. & Ovary & MCQ & — \\
Spatial morphology & Seeker & Clust. & Ovary & MCQ & — \\
Apoptosis trends & Seeker & Diff. Expr. & Ovary & Jaccard & J$\geq$0.5 \\
DE markers & Seeker & Diff. Expr. & Ovary & P@K & P@5 \\
Differential expression & Seeker & Diff. Expr. & Ovary & P@K & P@5 \\
Follicle markers & Seeker & Diff. Expr. & Ovary & P@K & P@5 \\
Follicle markers & Seeker & Diff. Expr. & Ovary & P@K & P@5 \\
Follicle markers & Seeker & Diff. Expr. & Ovary & P@K & P@5 \\
Dim. reduction & Seeker & Dim. Red. & Ovary & MCQ & — \\
PC gene interpretation & Seeker & Dim. Red. & Ovary & P@K & P@5 \\
PC loading interpretation & Seeker & Dim. Red. & Ovary & MCQ & — \\
PC loading interpretation & Seeker & Dim. Red. & Ovary & MCQ & — \\
PC pathway association & Seeker & Dim. Red. & Ovary & MCQ & — \\
PC population separation & Seeker & Dim. Red. & Ovary & MCQ & — \\
Count scaling & Seeker & Norm. & Ovary & Numeric & $\pm$0.1 \\
Count transformation & Seeker & Norm. & Ovary & Numeric & $\pm$0.1 \\
Background removal & Seeker & QC & Ovary & Numeric & $\pm$2000 \\
Min UMI filtering & Seeker & QC & Ovary & Numeric & $\pm$680 \\
Min genes filtering & Seeker & QC & Ovary & Numeric & $\pm$690 \\
Mitochondrial filtering & Seeker & QC & Ovary & Numeric & $\pm$582 \\
Follicle counting & Seeker & Spatial & Ovary & Numeric & $\pm$20 \\
LR directionality & Seeker & Spatial & Ovary & MCQ & — \\
LR temporal changes & Seeker & Spatial & Ovary & MCQ & — \\
Ligand-receptor analysis & Seeker & Spatial & Ovary & P@K & P@5 \\
Receptor family analysis & Seeker & Spatial & Ovary & MCQ & — \\
Signaling diversity & Seeker & Spatial & Ovary & MCQ & — \\
\midrule
Cell deconvolution & Visium & Cell Type & Bone & Jaccard & J$\geq$0.5 \\
Fibroblast ID & Visium & Cell Type & Bone & P@K & P@5 \\
Macrophage ID & Visium & Cell Type & Bone & P@K & P@5 \\
Monocyte ID & Visium & Cell Type & Bone & P@K & P@5 \\
NK cell ID & Visium & Cell Type & Bone & P@K & P@5 \\
NK cell ID & Visium & Cell Type & Bone & P@K & P@5 \\
Osteoclast ID & Visium & Cell Type & Bone & P@K & P@5 \\
RUNX2 correlation & Visium & Cell Type & Bone & Numeric & — \\
Cluster functions & Visium & Clust. & Bone & MCQ & — \\
Clustering & Visium & Clust. & Bone & P@K & P@5 \\
Niche cluster markers & Visium & Clust. & Bone & P@K & P@5 \\
Niche cluster markers & Visium & Clust. & Bone & P@K & P@5 \\
Spatial cluster coherence & Visium & Clust. & Bone & MCQ & — \\
DE gene markers & Visium & Diff. Expr. & Bone & P@K & P@5 \\
Enrichment analysis & Visium & Diff. Expr. & Bone & MCQ & — \\
Gradient ranking & Visium & Diff. Expr. & Bone & MCQ & — \\
Spatial gradient pathways & Visium & Diff. Expr. & Bone & MCQ & — \\
HVG selection & Visium & Dim. Red. & Bone & P@K & P@5 \\
UMAP interpretation & Visium & Dim. Red. & Bone & MCQ & — \\
UMAP neighbor analysis & Visium & Dim. Red. & Bone & MCQ & — \\
UMAP proximity & Visium & Dim. Red. & Bone & MCQ & — \\
Variance normalization & Visium & Norm. & Bone & Numeric & $\pm$1 \\
Min UMI filtering & Visium & QC & Bone & Numeric & $\pm$600 \\
Min genes filtering & Visium & QC & Bone & Numeric & $\pm$600 \\
QC filtering & Visium & QC & Bone & Numeric & $\pm$800 \\
Sparse gene filtering & Visium & QC & Bone & Numeric & $\pm$2000 \\
Stressed cell filtering & Visium & QC & Bone & Numeric & $\pm$200 \\
Ligand-receptor analysis & Visium & Spatial & Bone & Jaccard & J$\geq$0.5 \\
Niche pathways & Visium & Spatial & Bone & P@K & P@5 \\
Spatial dependency & Visium & Spatial & Bone & MCQ & — \\
Spatial niches & Visium & Spatial & Bone & Jaccard & J$\geq$0.5 \\
Spatial niches & Visium & Spatial & Bone & P@K & P@5 \\
\midrule
Cell type distribution & Xenium & Cell Type & Kidney & Cosine & cos$\geq$0.8 \\
Cell type distribution & Xenium & Cell Type & Kidney & Cosine & cos$\geq$0.8 \\
Cell type distribution & Xenium & Cell Type & Kidney & Numeric & — \\
Cell type distribution & Xenium & Cell Type & Kidney & Cosine & cos$\geq$0.8 \\
Cell type markers & Xenium & Cell Type & Kidney & P@K & P@5 \\
Interstitial typing & Xenium & Cell Type & Kidney & Jaccard & J$\geq$0.5 \\
Cluster enrichment & Xenium & Clust. & Kidney & Numeric & $\pm$0.2 \\
Injury niche clustering & Xenium & Clust. & Kidney & MCQ & — \\
Interstitial clustering & Xenium & Clust. & Kidney & MCQ & — \\
Niche cluster markers & Xenium & Clust. & Kidney & P@K & P@5 \\
Cell type changes & Xenium & Diff. Expr. & Kidney & Jaccard & J$\geq$0.5 \\
Cell type changes & Xenium & Diff. Expr. & Kidney & Jaccard & J$\geq$0.5 \\
DE gene identification & Xenium & Diff. Expr. & Kidney & Jaccard & J$\geq$0.5 \\
DE gene identification & Xenium & Diff. Expr. & Kidney & P@K & P@5 \\
ECM pathway analysis & Xenium & Diff. Expr. & Kidney & MCQ & — \\
Pathway analysis & Xenium & Diff. Expr. & Kidney & MCQ & — \\
Signaling programs & Xenium & Diff. Expr. & Kidney & MCQ & — \\
PC loading interpretation & Xenium & Dim. Red. & Kidney & MCQ & — \\
PC module pairing & Xenium & Dim. Red. & Kidney & MCQ & — \\
PCA cell separation & Xenium & Dim. Red. & Kidney & MCQ & — \\
Log1p normalization & Xenium & Norm. & Kidney & Numeric & $\pm$0.1 \\
Normalization & Xenium & Norm. & Kidney & Numeric & $\pm$0.1 \\
Detection summary & Xenium & QC & — & Numeric & $\pm$20 \\
Min UMI filtering & Xenium & QC & — & Numeric & $\pm$50 \\
Min genes filtering & Xenium & QC & — & Numeric & $\pm$50 \\
Colocalization patterns & Xenium & Spatial & Kidney & MCQ & — \\
Neighborhood dynamics & Xenium & Spatial & Kidney & MCQ & — \\
Niche identity & Xenium & Spatial & Kidney & MCQ & — \\
Spatial analysis & Xenium & Spatial & Kidney & MCQ & — \\
Spatial signaling & Xenium & Spatial & Kidney & MCQ & — \\
\end{longtable}

\section*{A.2 Canonical Problem Examples}

This section presents 10 canonical examples from SpatialBench that illustrate the diversity of tasks, platforms, and grader types in the benchmark. These examples are included in the public release to demonstrate the evaluation format without exposing the full benchmark. Each example includes the task prompt, platform, grader type, and expected answer format.

\subsection*{Overview}

Table~\ref{tab:canonical-overview} summarizes the 10 canonical examples.

\begin{table}[h]
\centering
\caption{Overview of canonical examples included in SpatialBench.}
\label{tab:canonical-overview}
\small
\begin{tabular}{clllll}
\toprule
\# & Task Category & Platform & Tissue & Grader & Key Parameter \\
\midrule
1 & Quality Control & Xenium & Kidney & Numeric & $\pm$50 cells \\
2 & Normalization & MERFISH & Brain & Numeric & $\pm$0.05 \\
3 & Dim. Reduction & Seeker & Ovary & MCQ & — \\
4 & Clustering & Visium & Bone & P@K & P@5 \\
5 & Clustering & MERFISH & Brain & MCQ & — \\
6 & Cell Typing & Xenium & Kidney & Cosine & $\geq$0.8 \\
7 & Cell Typing & MERFISH & Brain & P@K & P@5 \\
8 & Diff. Expression & Seeker & Ovary & P@K & P@5 \\
9 & Spatial Analysis & Visium & Bone & Jaccard & $\geq$0.5 \\
10 & Spatial Analysis & Xenium & Kidney & MCQ & — \\
\bottomrule
\end{tabular}
\end{table}

\subsection*{Example 1: Quality Control (Xenium)}

\textbf{ID:} \texttt{xenium\_qc\_filter\_min\_umi\_counts}

\textbf{Task:} Filter cells with very low total UMI counts. Return a JSON object with field: \texttt{cells\_after\_filtering} (number remaining after filter).

\textbf{Grader:} NumericTolerance with absolute tolerance of $\pm$50 cells.

\textbf{Rationale:} Tests whether the agent can apply standard quality control thresholds appropriate for spatial transcriptomics data. Spatial platforms typically have lower UMI counts per cell than dissociated single-cell RNA-seq, requiring calibrated filtering thresholds.

\subsection*{Example 2: Normalization (MERFISH)}

\textbf{ID:} \texttt{merfish\_brain\_log\_zscore\_gad2\_mean}

\textbf{Task:} Using the raw MERFISH brain dataset, first normalize per cell by scaling so that the sum of gene expression values per cell equals 250. Then log-transform the scaled values using log1p. Next, Z-score each gene independently across cells. After z-scoring, compute the mean Z-scored expression of gene `Gad2' across all cells. Return: \texttt{\{"mean\_gene\_value": <float>\}}.

\textbf{Grader:} NumericTolerance with absolute tolerance of $\pm$0.05.

\textbf{Rationale:} Tests multi-step normalization pipeline execution. The expected answer (mean z-score $\approx$ 0) validates correct implementation of the z-scoring step, which by definition produces zero-mean values per gene.

\subsection*{Example 3: Dimensionality Reduction (Seeker)}

\textbf{ID:} \texttt{seeker\_3x3\_ovary\_1hr\_pc1\_cell\_populations}

\textbf{Task:} PC1 primarily separates which two cell populations?
\begin{itemize}
    \item[A)] Oocytes vs Granulosa cells
    \item[B)] Granulosa cells vs Theca/Stroma cells
    \item[C)] Theca cells vs Stromal fibroblasts
    \item[D)] Endothelial cells vs Smooth muscle
\end{itemize}
Return: \texttt{\{"answer": "<letter>"\}}.

\textbf{Grader:} MultipleChoice.

\textbf{Rationale:} Tests interpretation of PCA results in biological context. The agent must perform PCA, examine loadings or cell distributions along PC1, and identify which cell populations are separated by the dominant axis of variation.

\subsection*{Example 4: Clustering (Visium)}

\textbf{ID:} \texttt{bone\_visium\_c2l\_niche\_bone\_formation\_markers}

\textbf{Task:} Attached is a human bone spatial dataset with estimated cell-type abundance and gene expression. Cluster spots to identify niches, then find the bone-formation niche (active osteogenesis) and return its top 10 enriched marker genes; if none is identified, return an empty list. Return: \texttt{\{"top\_marker\_genes": ["..."]\}}.

\textbf{Grader:} MarkerGenePrecisionRecall at P@5.

\textbf{Rationale:} Tests spatial niche identification and marker gene discovery. The agent must cluster Visium spots based on cell-type composition, identify the biologically relevant niche, and extract differentially expressed markers.

\subsection*{Example 5: Clustering (MERFISH)}

\textbf{ID:} \texttt{merfish\_brain\_clustering\_astro2\_vs\_astro}

\textbf{Task:} Using filtered, normalized MERFISH brain data, perform clustering based solely on expression data. Compare how Astro-2 cells distribute relative to other astrocyte states. Which pattern best matches?
\begin{itemize}
    \item[A)] Astro-2 cells are evenly scattered across multiple astrocyte clusters
    \item[B)] Astro-2 cells are enriched in a single dominant cluster transcriptionally distinct from Astro-1
    \item[C)] Astro-2 cells form their own isolated cluster with no overlap
    \item[D)] Astro-2 cells cluster with non-astrocyte cell types
\end{itemize}
Return: \texttt{\{"answer": "<letter>"\}}.

\textbf{Grader:} MultipleChoice.

\textbf{Rationale:} Tests clustering analysis and biological interpretation. The agent must cluster cells, cross-reference with known cell type annotations, and evaluate cluster composition to determine the correct biological pattern.

\subsection*{Example 6: Cell Typing (Xenium)}

\textbf{ID:} \texttt{classify\_pt\_distribution\_advanced}

\textbf{Task:} Infer PT subtypes from gene expression (marker-based) and assign each cell exactly one label from: PTS1 (S1 segment), PTS2 (S2 segment), PTS3 (S3 segment), Inj-PT (injured), FR-PT (failed repair). Report the subtype distribution as percentages of total PT cells. Return: \texttt{\{"total\_cells": <int>, "cell\_type\_distribution": \{...\}\}}.

\textbf{Grader:} DistributionComparison with cosine similarity threshold $\geq$0.8.

\textbf{Rationale:} Tests marker-based cell type classification in disease context. The agent must use known PT segment markers and injury-associated genes to classify cells and report accurate population proportions in a kidney injury model.

\subsection*{Example 7: Cell Typing (MERFISH)}

\textbf{ID:} \texttt{merfish\_brain\_astrocyte\_aging\_markers}

\textbf{Task:} Using MERFISH brain data without cell type labels, identify astrocytes and distinguish an aging/inflammation-associated astrocyte state from baseline astrocyte states based on expression and clustering. Select a small set of significantly enriched marker genes for the aging-associated state. Return: \texttt{\{"top\_marker\_genes": ["..."]\}}.

\textbf{Grader:} MarkerGenePrecisionRecall at P@5.

\textbf{Rationale:} Tests de novo cell type identification and state-specific marker discovery. The agent must first identify astrocytes without labels, then find a biologically meaningful substate and its characteristic markers.

\subsection*{Example 8: Differential Expression (Seeker)}

\textbf{ID:} \texttt{ovary\_mural\_gc\_atretic\_follicle\_markers}

\textbf{Task:} Using mouse ovary spatial data with cell type annotations, find the top 20 differentially expressed genes in Atretic follicle mural granulosa cells compared to other follicle types (Antral, Preantral). Subset to mural granulosa cells from Immature ovary samples only. Exclude ribosomal genes (Rps*, Rpl*, Mrp*) from results. Return: \texttt{\{"top\_marker\_genes": ["..."]\}}.

\textbf{Grader:} MarkerGenePrecisionRecall at P@5.

\textbf{Rationale:} Tests conditional differential expression analysis. The agent must correctly subset data based on multiple criteria, perform DE analysis, and filter results appropriately.

\subsection*{Example 9: Spatial Analysis (Visium)}

\textbf{ID:} \texttt{visium\_spatial\_niches\_bone\_meslin}

\textbf{Task:} Using human bone Visium data with cell2location deconvolution results, identify exactly 2 spatial niches using clustering. Determine which cell types are enriched in the osteogenic niche. Return: \texttt{\{"osteogenic\_enriched\_celltypes": [...]\}}.

\textbf{Grader:} JaccardLabelSet with threshold $\geq$0.5.

\textbf{Rationale:} Tests spatial niche identification and cell type enrichment analysis. The agent must integrate deconvolution results with spatial clustering to identify biologically meaningful tissue compartments.

\subsection*{Example 10: Spatial Analysis (Xenium)}

\textbf{ID:} \texttt{xenium\_kidney\_cn3\_pts3\_neighborhood\_dynamics}

\textbf{Task:} Based on annotated cell types, how does the fraction of cells assigned to CN3 change over the ischemia–reperfusion injury time course?
\begin{itemize}
    \item[A)] Present at all timepoints with stable abundance
    \item[B)] Absent during mid-injury phase (hour 12 and day 2) and reappears during repair
    \item[C)] Present only at 6 weeks post-injury
    \item[D)] Restricted to the medulla at all timepoints with no temporal change
\end{itemize}
Return: \texttt{\{"answer": "<letter>"\}}.

\textbf{Grader:} MultipleChoice.

\textbf{Rationale:} Tests temporal dynamics analysis in spatial context. The agent must track cell neighborhood composition changes across injury and repair phases, requiring integration of spatial annotations with temporal metadata.

\subsection*{Evaluation Format}

All evaluations follow a standardized JSON format:

\begin{verbatim}
{
  "id": "evaluation_identifier",
  "task": "Natural language task description...",
  "data_node": "latch://path/to/dataset.h5ad",
  "grader": {
    "type": "grader_type",
    "config": {
      "ground_truth": {...},
      "tolerances": {...}
    }
  },
  "metadata": {
    "task": "task_category",
    "kit": "platform_name"
  }
}
\end{verbatim}

The canonical examples are available in the \texttt{evals\_canonical/} directory of the SpatialBench repository, organized by task category.

\section*{A.3 Grader Specification}

This section documents the grader system used in SpatialBench, including grader families, tolerance rules, answer normalization, and validation methodology.

\subsection*{Overview}

SpatialBench implements a modular grading system with 5 primary grader families covering different answer types encountered in spatial biology analysis. Each grader inherits from a base class that handles answer extraction from agent responses and delegates evaluation to type-specific logic.

\begin{table}[h]
\centering
\caption{Distribution of grader types across SpatialBench evaluations.}
\label{tab:grader-distribution}
\small
\begin{tabular}{lcc}
\toprule
Grader Family & Evaluations & Primary Use Cases \\
\midrule
MultipleChoiceGrader & 45 & Interpretation, biological pattern identification \\
MarkerGenePrecisionRecallGrader & 45 & Marker discovery, differential expression \\
NumericToleranceGrader & 36 & QC metrics, counts, expression values \\
LabelSetJaccardGrader & 18 & Cell type prediction, niche composition \\
DistributionComparisonGrader & 3 & Cell population proportions \\
\bottomrule
\end{tabular}
\end{table}

\subsection*{Grader Families}

\subsubsection*{1. NumericToleranceGrader}

Evaluates numeric answers against ground truth values with configurable tolerance.

\textbf{Input format:}
\begin{verbatim}
{"field_name": <numeric_value>}
\end{verbatim}

\textbf{Tolerance types:}
\begin{itemize}
    \item \texttt{absolute}: Pass if $|x_{\text{pred}} - x_{\text{true}}| \leq \epsilon$
    \item \texttt{relative}: Pass if $\frac{|x_{\text{pred}} - x_{\text{true}}|}{|x_{\text{true}}|} \leq \epsilon$
    \item \texttt{min}: Pass if $x_{\text{pred}} \geq x_{\text{true}}$ (lower bound)
    \item \texttt{max}: Pass if $x_{\text{pred}} \leq x_{\text{true}}$ (upper bound)
\end{itemize}

\textbf{Multi-field evaluation:} When multiple fields are specified, all must pass for overall success.

\textbf{Example configuration:}
\begin{verbatim}
{
  "type": "numeric_tolerance",
  "config": {
    "ground_truth": {"cells_after_filtering": 1374915},
    "tolerances": {
      "cells_after_filtering": {"type": "absolute", "value": 50}
    }
  }
}
\end{verbatim}

\textbf{Tolerance rationale:} Absolute tolerances are calibrated to the expected variance from reasonable implementation choices. For cell counts, tolerances of 50--100 cells accommodate minor differences in filtering order or edge cases. For normalized expression values (e.g., z-scores), tolerances of 0.05 account for floating-point precision and minor algorithmic differences.

\subsubsection*{2. MultipleChoiceGrader}

Evaluates discrete answer selection from predefined options.

\textbf{Input format:}
\begin{verbatim}
{"answer": "<letter>"}
\end{verbatim}

\textbf{Matching rules:}
\begin{itemize}
    \item Case-insensitive comparison (``a'', ``A'', and `` A '' all match)
    \item Whitespace trimming applied before comparison
    \item Exact single-character match required
\end{itemize}

\textbf{Example configuration:}
\begin{verbatim}
{
  "type": "multiple_choice",
  "config": {
    "correct_answer": "B"
  }
}
\end{verbatim}

\textbf{Design rationale:} Multiple-choice questions test biological interpretation rather than exact computation. Each question includes 3--4 plausible distractors designed to capture common misconceptions or analysis errors.

\subsubsection*{3. MarkerGenePrecisionRecallGrader}

Evaluates marker gene lists using precision and recall at K.

\textbf{Input format:}
\begin{verbatim}
{"top_marker_genes": ["Gene1", "Gene2", ...]}
\end{verbatim}

\textbf{Metrics:}
\begin{align}
\text{Precision@K} &= \frac{|\text{Predicted} \cap \text{Canonical}|}{K} \\
\text{Recall@K} &= \frac{|\text{Predicted} \cap \text{Canonical}|}{|\text{Canonical}|}
\end{align}

\textbf{Matching rules:}
\begin{itemize}
    \item Case-insensitive gene name comparison (\texttt{COL1A1} matches \texttt{col1a1})
    \item No synonym expansion (gene symbols must match exactly after case normalization)
    \item Set-based comparison (order within the predicted list does not affect scoring)
\end{itemize}

\textbf{Default thresholds:} Precision@K $\geq$ 0.60, Recall@K $\geq$ 0.50

\textbf{Example configuration:}
\begin{verbatim}
{
  "type": "marker_gene_precision_recall",
  "config": {
    "canonical_markers": ["COL1A1", "COL1A2", "SPP1", "SPARC", "BGLAP", "IBSP"],
    "scoring": {
      "pass_thresholds": {
        "precision_at_k": 0.0,
        "recall_at_k": 0.5
      }
    }
  }
}
\end{verbatim}

\textbf{Threshold rationale:} Recall thresholds are set based on the number of canonical markers and biological redundancy. For small canonical sets (5--10 genes), recall $\geq$ 0.5 ensures that the agent identifies the core biological signal. Precision thresholds are often relaxed (0.0--0.3) to allow discovery of valid but unlisted markers.

\subsubsection*{4. LabelSetJaccardGrader}

Evaluates predicted label sets against ground truth using Jaccard similarity.

\textbf{Input format:}
\begin{verbatim}
{"cell_types_predicted": ["Type1", "Type2", ...]}
\end{verbatim}

\textbf{Metric:}
\begin{equation}
J(A, B) = \frac{|A \cap B|}{|A \cup B|}
\end{equation}

\textbf{Matching rules:}
\begin{itemize}
    \item Exact string matching (case-sensitive)
    \item No partial matching or synonym expansion
    \item Both false positives and false negatives penalized symmetrically
\end{itemize}

\textbf{Default threshold:} Jaccard $\geq$ 0.90

\textbf{Example configuration:}
\begin{verbatim}
{
  "type": "jaccard_label_set",
  "config": {
    "ground_truth_labels": ["Mesenchymal lineage"],
    "scoring": {
      "pass_threshold": 1.0
    }
  }
}
\end{verbatim}

\textbf{Threshold rationale:} High thresholds (0.9--1.0) are used when the ground truth set is small and unambiguous. Lower thresholds (0.5--0.7) accommodate tasks where multiple valid label sets exist.

\subsubsection*{5. DistributionComparisonGrader}

Evaluates predicted cell type distributions against ground truth proportions.

\textbf{Input format:}
\begin{verbatim}
{
  "total_cells": <integer>,
  "cell_type_distribution": {
    "Type1": <percentage>,
    "Type2": <percentage>,
    ...
  }
}
\end{verbatim}

\textbf{Matching rules:}
\begin{itemize}
    \item Per-category absolute tolerance on percentages (typically $\pm$3--5\%)
    \item Optional total cell count validation
    \item Missing categories treated as failures
    \item Extra categories (not in ground truth) flagged but may not fail depending on configuration
\end{itemize}

\textbf{Example configuration:}
\begin{verbatim}
{
  "type": "distribution_comparison",
  "config": {
    "ground_truth": {
      "cell_type_distribution": {
        "Inj_PT": 48.55, "PTS2": 5.02, "PTS1": 42.06,
        "PTS3": 0.9, "FR_PT": 3.47
      }
    },
    "tolerances": {
      "cell_type_percentages": {"type": "absolute", "value": 5.0}
    }
  }
}
\end{verbatim}

\textbf{Tolerance rationale:} Percentage tolerances of $\pm$5\% accommodate variation from different marker thresholds and assignment algorithms while ensuring the agent captures the correct rank ordering and approximate proportions.

\subsection*{Answer Extraction}

All graders use a common answer extraction mechanism from agent responses:

\begin{enumerate}
    \item Scan assistant messages in reverse chronological order
    \item Locate \texttt{submit\_response} tool calls
    \item Extract JSON from \texttt{<EVAL\_ANSWER>...</EVAL\_ANSWER>} tags
    \item Parse JSON and validate required fields
\end{enumerate}

If answer extraction fails (no tags found, malformed JSON, or missing fields), the evaluation automatically fails with an explanatory message.

\subsection*{Tolerance Design Principles}

\subsubsection*{Numeric Tolerances}

Numeric tolerances are calibrated based on:

\begin{enumerate}
    \item \textbf{Biological variance}: Expected variation from legitimate methodological choices (e.g., exact filtering threshold)
    \item \textbf{Computational precision}: Floating-point differences from library versions or implementation details
    \item \textbf{Discrimination}: Tight enough to distinguish correct from incorrect approaches
\end{enumerate}

For example, the QC cell count tolerance of $\pm$50 cells on a dataset of 1.4M cells ($<$0.004\%) is calibrated to:
\begin{itemize}
    \item Accept minor threshold variations (e.g., $<$10 vs $\leq$10 UMIs)
    \item Reject fundamentally wrong approaches (e.g., filtering 40\% of cells)
\end{itemize}

\subsubsection*{Gene Name Normalization}

Gene names are normalized to lowercase before comparison. This handles:
\begin{itemize}
    \item Species conventions (\texttt{COL1A1} in human vs \texttt{Col1a1} in mouse)
    \item Database variations (HGNC vs Ensembl symbol formatting)
    \item Agent output variations
\end{itemize}

Synonym expansion (e.g., \texttt{CD4} $\leftrightarrow$ \texttt{Cd4 antigen}) is deliberately not implemented to:
\begin{itemize}
    \item Maintain reproducibility across synonym database versions
    \item Avoid ambiguous many-to-many mappings
    \item Encourage agents to use canonical nomenclature from the dataset
\end{itemize}

\subsubsection*{Multiple-Choice Normalization}

Answer normalization includes:
\begin{itemize}
    \item Uppercase conversion
    \item Whitespace trimming
    \item Single-character extraction
\end{itemize}

This accepts variations like \texttt{``b''}, \texttt{``B''}, \texttt{`` B ''}, and \texttt{``B)''} as equivalent.

\subsection*{Validation and Quality Control}

Grader correctness and robustness were validated through multiple mechanisms:

\subsubsection*{1. Ground Truth Verification}

Each evaluation's ground truth was independently computed by:
\begin{enumerate}
    \item Manual analysis following the task specification
    \item Cross-validation against published results where available
    \item Verification by a second analyst for complex evaluations
\end{enumerate}

\subsubsection*{2. Tolerance Calibration}

Tolerances were calibrated by:
\begin{enumerate}
    \item Computing ground truth with multiple valid approaches
    \item Measuring the range of correct answers
    \item Setting tolerances to accept all valid approaches while rejecting incorrect ones
    \item Testing with adversarial edge cases (e.g., off-by-one errors, wrong gene)
\end{enumerate}

\subsubsection*{3. Grader Unit Testing}

Each grader includes test coverage for:
\begin{itemize}
    \item Exact match (pass)
    \item Within tolerance (pass)
    \item Outside tolerance (fail)
    \item Missing fields (fail with message)
    \item Type errors (fail gracefully)
    \item Edge cases (empty lists, zero values, special characters)
\end{itemize}

\subsubsection*{4. Anti-Gaming Measures}

Evaluations are designed to resist trivial exploitation:
\begin{itemize}
    \item Numeric answers require actual computation (cannot be guessed from task wording)
    \item Gene lists require analysis of the specific dataset (canonical markers vary by context)
    \item Multiple-choice distractors are biologically plausible
    \item Distribution tasks require correct identification of all categories
\end{itemize}

\subsubsection*{5. Manual Trajectory Review}

A random sample of agent trajectories was manually reviewed to verify that:
\begin{itemize}
    \item Passing evaluations reflect genuine understanding
    \item Failing evaluations identify real errors
    \item Edge cases are handled appropriately
\end{itemize}

\subsection*{Grader Configuration Schema}

All graders follow a consistent JSON schema:

\begin{verbatim}
{
  "type": "<grader_type>",
  "config": {
    "ground_truth": {...},      // Expected answer(s)
    "tolerances": {...},        // Type-specific tolerance config
    "scoring": {...}            // Thresholds and pass criteria
  }
}
\end{verbatim}

Valid grader types:
\begin{itemize}
    \item \texttt{numeric\_tolerance}
    \item \texttt{multiple\_choice}
    \item \texttt{marker\_gene\_precision\_recall}
    \item \texttt{jaccard\_label\_set}
    \item \texttt{distribution\_comparison}
\end{itemize}

\subsection*{Extensibility}

New graders can be added by:
\begin{enumerate}
    \item Subclassing \texttt{BinaryGrader}
    \item Implementing the \texttt{evaluate\_answer(agent\_answer, config)} method
    \item Registering in \texttt{GRADER\_REGISTRY}
\end{enumerate}

The grader system is designed to be framework-agnostic, accepting any JSON-serializable answer format.

\section*{A.4 Statistical Methods}

This section details the statistical procedures used for computing confidence intervals, handling missing data, and aggregating results across models and evaluation categories.

\subsection*{Experimental Design}

Each model configuration was evaluated on 146 evaluations across 3 independent runs. Runs were executed with identical evaluation sets but independent random seeds for any stochastic agent behavior. The total sample comprises 9 model configurations $\times$ 3 runs $\times$ 146 evaluations = 3,942 individual evaluation attempts.

\subsection*{Confidence Interval Procedure}

\subsubsection*{Two-Stage Aggregation}

Confidence intervals are computed using a two-stage aggregation procedure:

\textbf{Stage 1: Per-evaluation averaging.} For each evaluation $e$, compute the mean score across runs:
\begin{equation}
\bar{s}_e = \frac{1}{R} \sum_{r=1}^{R} s_{e,r}
\end{equation}
where $s_{e,r} \in \{0, 1\}$ is the pass/fail indicator for evaluation $e$ in run $r$, and $R$ is the number of runs (typically 3).

\textbf{Stage 2: Cross-evaluation inference.} Treat the $N$ per-evaluation means $\{\bar{s}_1, \ldots, \bar{s}_N\}$ as the sample for inference. Compute the overall accuracy as:
\begin{equation}
\hat{\mu} = \frac{1}{N} \sum_{e=1}^{N} \bar{s}_e
\end{equation}

\subsubsection*{Confidence Interval Construction}

We use a t-distribution-based interval to account for uncertainty in the variance estimate:

\begin{equation}
\hat{\sigma}^2 = \frac{1}{N-1} \sum_{e=1}^{N} (\bar{s}_e - \hat{\mu})^2
\end{equation}

\begin{equation}
\text{SE} = \sqrt{\frac{\hat{\sigma}^2}{N}}
\end{equation}

\begin{equation}
\text{CI}_{95\%} = \hat{\mu} \pm t_{0.975, N-1} \times \text{SE}
\end{equation}

where $t_{0.975, N-1}$ is the 97.5th percentile of the Student's t-distribution with $N-1$ degrees of freedom.

\textbf{Rationale:} This procedure treats evaluations as the unit of analysis, acknowledging that evaluations vary in difficulty and that the benchmark's coverage of the evaluation space is the primary source of uncertainty. CIs computed over runs would underestimate uncertainty by ignoring evaluation-level variance.

\subsubsection*{Subgroup Analysis}

For task-specific and platform-specific accuracy, the same procedure is applied within each subgroup:
\begin{itemize}
    \item Task categories: QC, Normalization, Dimensionality Reduction, Clustering, Cell Typing, Differential Expression, Spatial Analysis
    \item Platforms: Xenium, Visium, MERFISH, Seeker, AtlasXomics
\end{itemize}

Subgroup CIs have larger widths due to smaller sample sizes (e.g., $N=7$ for Normalization vs $N=39$ for Cell Typing).

\subsection*{Efficiency Metrics}

The same two-stage procedure is applied to efficiency metrics:

\textbf{Steps per evaluation:}
\begin{equation}
\bar{k}_e = \frac{1}{R} \sum_{r=1}^{R} k_{e,r}
\end{equation}
where $k_{e,r}$ is the number of agent steps for evaluation $e$ in run $r$.

\textbf{Duration and cost:} Analogous formulas for $\bar{t}_e$ (seconds) and $\bar{c}_e$ (USD).

\subsection*{Handling of Missing Data}

\subsubsection*{Timeouts and Crashes}

Evaluations that exceed the step limit (100 steps) or crash are recorded with:
\begin{itemize}
    \item \texttt{passed}: \texttt{false}
    \item \texttt{missing}: \texttt{true} (for crashed evaluations only)
    \item Efficiency metrics: recorded up to the point of timeout/crash
\end{itemize}

Missing evaluations are included in accuracy calculations as failures. This is conservative: it penalizes models that crash rather than treating missing data as ignorable.

\subsubsection*{Reference-Based Imputation}

When aggregating across runs, missing evaluations are imputed from a reference run:
\begin{enumerate}
    \item Identify the set of expected evaluations from a complete reference run
    \item For each missing evaluation, insert a failure record
    \item Proceed with standard aggregation
\end{enumerate}

This ensures consistent evaluation counts across models and runs.

\subsection*{Bounded Metrics}

\subsubsection*{Accuracy Bounds}

Accuracy is naturally bounded in $[0, 1]$. Confidence intervals computed via the t-interval formula may extend below 0 or above 1 when accuracy is near the boundaries and variance is high. For presentation, reported intervals are clipped to $[0, 100]\%$.

\subsubsection*{Step Count Bounds}

Step counts are bounded below by 1 (minimum one step to attempt the task) and above by 100 (the step limit). The step limit is a hard constraint; evaluations reaching 100 steps are recorded as 100 regardless of whether additional steps would have been attempted.

\textbf{Clipping rationale:} The step limit represents a practical constraint on agent computation budget, not a true upper bound on difficulty. We report step counts as-is without adjustment.

\subsubsection*{Cost Bounds}

Cost is bounded below by 0 and has no explicit upper bound. However, cost correlates with step count, so the step limit implicitly bounds cost per evaluation. A small number of evaluations ($<$2\%) had missing cost data due to API logging issues; these were excluded from cost aggregates but included in accuracy calculations.

\subsection*{Multiple Comparisons}

We do not apply multiple comparison corrections to the reported confidence intervals for several reasons:

\begin{enumerate}
    \item \textbf{Exploratory analysis:} The benchmark is intended for model comparison and capability assessment, not hypothesis testing.
    \item \textbf{Dependent structure:} Evaluations within task categories share systematic difficulty, violating independence assumptions of standard corrections.
    \item \textbf{Interpretability:} Uncorrected 95\% CIs are directly interpretable as reflecting sampling uncertainty.
\end{enumerate}

Readers should interpret overlapping CIs as indicating plausible equivalence, not definitive equality.

\subsection*{Tie-Breaking}

For model rankings (e.g., ``best model by task''), ties are broken by:
\begin{enumerate}
    \item Point estimate (higher is better for accuracy)
    \item Confidence interval width (narrower is preferred)
    \item Alphabetical order (as a last resort)
\end{enumerate}

In practice, point estimates differ sufficiently that tie-breaking is rarely needed.

\end{document}